\pdfoutput=1
\documentclass[twoside]{article}

\usepackage[preprint]{aistats2027}

\usepackage[round]{natbib}

\usepackage{hyperref}
\usepackage{url}

\usepackage{amsfonts}
\usepackage{mathtools}
\usepackage{nicefrac}
\usepackage{dsfont}

\usepackage{graphicx}
\usepackage{caption}
\usepackage{subcaption}

\usepackage{booktabs}

\usepackage{algorithm}
\usepackage{algorithmic}

\usepackage{tikz}
\usetikzlibrary{bayesnet}

\usepackage{comment}

\newcommand{\iid}{i.i.d. }
\newcommand{\ie}{i.e., }

\newcommand{\eg}{e.g., }

\newcommand{\Real}{{\mathbb R}}

\newcommand{\eValue}[2]{\mathbb{E}_{#1}\left\{ #2 \right\}}

\newcommand{\N}[1]{\mathcal{N}\left( #1\right)}

\newcommand{\T}[1]{t\left( #1\right)}
\newcommand{\MT}[1]{\mathcal{T}\left( #1\right)}

\newcommand{\IG}[1]{\mathcal{IG}\left( #1\right)}

\newcommand{\U}[1]{\mathcal{U}\left( #1\right)}

\newcommand{\tr}[1]{\mathrm{tr}\left\{ #1 \right\}} %

\newcommand{\dd}[1]{\mathrm{d} #1}

\newcommand{\A}{\mathcal{A}}

\newcommand{\ptilde}{\tilde{p}}

\newcommand{\thetastar}{\theta^{*}}

\newcommand{\varphitilde}{\tilde{\varphi}}

\newcommand{\HH}{\mathcal{H}}

\newcommand{\argmax}{\mathop{\mathrm{argmax}}}

\begin{document}

\runningtitle{Bayesian Nonparametric Thompson Sampling for Multi-Armed Bandits}

\twocolumn[

\aistatstitle{
Bayesian Nonparametric Thompson Sampling \\ 
for Multi-Armed Bandits
}

\aistatsauthor{ I\~{n}igo Urteaga \And Chris Wiggins}

\aistatsaddress{ BCAM -- Basque Center for Applied Mathematics \\ Ikerbasque -- Basque Foundation for Science \And  Columbia University} ]

\begin{abstract}
In stochastic multi-armed bandits, rewards are generated by an unknown distribution; the world's reward uncertainty induces the exploration-exploitation trade-off.
Thompson sampling (TS) addresses this trade-off elegantly, but requires knowing the reward model's parametric form to sample from its posterior and compute per-arm expected rewards ---which limits its applicability as it demands per-scenario model specification and hyperparameter tuning.
We here adopt Bayesian nonparametric (BNP) mixture models to remove this requirement, extending TS to arbitrary reward distributions.
We model each arm's unknown reward distribution with an independent BNP Gaussian mixture, which can approximate any continuous reward density, and adapts its complexity to sequentially observed data. Embedding the BNP mixture model into TS yields \texttt{Nonparametric TS}, an algorithm that sequentially learns the reward model that best approximates the true, yet unknown, per-arm reward distribution, avoiding case-by-case model specification and hyperparameter tuning.
We empirically demonstrate that
($i$) \texttt{Nonparametric TS} matches the regret performance of Oracle TS policies that know the true reward model;
($ii$) it favorably compares (in cumulative regret and volatility) to a suite of state-of-the-art alternatives across multimodal, outlier-contaminated, and non-exponential-family reward distributions;
and ($iii$) its Bayesian nonparametric inference remains computationally practical and accurate for regret minimization.
\end{abstract}

\section{Introduction}
\label{sec:intro}
Sequential decision making aims to optimize interactions with the world (exploit), while simultaneously learning how the world operates (explore).
The multi-armed bandit (MAB)~\citep{b-Lattimore2020} is a natural abstraction for a wide variety of real-world challenges that require learning while simultaneously maximizing rewards~\citep{j-Lai1985}, with applications spanning science, medicine, engineering and operations research~\citep{ip-Li2010,j-Villar2015,ip-Hill2017,j-Aziz2021}.
In practice, bandit applications are often contextual,
with continuous rewards (\eg dwell time in online advertising, or drug response in clinical trials).

Among the existing bandit algorithms,
Thompson sampling (TS)~\citep{j-Thompson1933} elegantly addresses the exploration-exploitation dilemma:
it updates a posterior over expected rewards for each arm, and chooses actions in proportion to their probability of being optimal, see Section~\ref{sec:background}.
TS, however, requires access to posterior samples of the \emph{true} reward model of the world, which is analytically tractable only for reward distributions in the exponential family~\citep{j-Russo2018}.
In practice, knowledge of the true reward model is illusory,
and balancing exploration and exploitation for reward distributions outside the exponential family remains largely unsolved.

Researchers have pursued this aim via Bayesian neural networks with approximate inference~\citep{ip-Riquelme2018},
bootstrapping~\citep{j-Osband2015},
ensembling~\citep{ip-Lu2017},
and Dirichlet-weighted re-sampling of observed rewards~\citep{ip-Riou2020,dirichletts2021}.
In contrast, we exploit Bayesian nonparametric (BNP) mixture models %
to perform MAB optimization under reward model uncertainty.
Unlike nonparametric priors over
reward distribution functions~\citep{dpps2025}
or over the MAB action space~\citep{ip-Srinivas2010,j-Battiston2018},
we focus on MABs with a discrete set of actions,
for which there is uncertainty about the \emph{per-arm reward distribution}.

The motivation for our proposal is a well-established result in Bayesian nonparametric statistics: under mild regularity conditions, the posterior of Gaussian mixtures under a Dirichlet process (DP) prior~\citep{j-Ferguson1973} converges to the true data-generating density as more observations are gathered, for a wide class of continuous densities~\citep{j-Ghosal1999} (see Section~\ref{sec:background}).
This is precisely the flexibility a Thompson sampling agent needs when the true reward distribution is unknown: rather than committing to a parametric family that may be misspecified,
we let a per-arm BNP mixture model grow in complexity as needed to explain observed rewards, and use its posterior predictive distribution to guide arm selection.

It is both the combination of BNP mixture models per-arm with Thompson sampling
and the resulting flexibility,
\ie avoiding model misspecification without case-by-case reward model design,
that is novel and significant in this work.
To the best of our knowledge, no other work uses BNP mixtures to model independent, per-arm MAB reward distributions.
In summary, our contributions are:
\begin{enumerate}
\item To present \texttt{Nonparametric TS}, a Thompson sampling algorithm in which each arm's reward distribution is modeled by an independent Dirichlet process (DP) Gaussian mixture model, grounded on established BNP posterior-convergence guarantees.
\texttt{Nonparametric TS}'s posterior predictive model adapts online to the observed rewards.
We additionally present natural extensions of the model in Appendix~\ref{asec:nonparametric_ts_extensions} to Pitman-Yor process priors (\ref{asec:pitman_yor}) and to a hierarchical BNP model (\ref{asec:hierarchical}).

\item To empirically demonstrate that \texttt{Nonparametric TS} matches the regret performance of an (unrealistic) Oracle TS that knows the true reward model. Performance is satisfactory across contextual linear, multimodal and outlier-contaminated bandits, and under model misspecification.

\item To empirically show that \texttt{Nonparametric TS} substantially reduces both average cumulative regret and regret volatility relative to a suite of state-of-the-art Thompson sampling alternatives.
\texttt{Nonparametric TS} dominates across bandits with multimodal, outlier-contaminated and exponentially distributed rewards, without any per-scenario tuning.

\item We show that the flexibility of Bayesian nonparametric posterior modeling comes at a practical computational cost: a warm-started, bounded-iteration Gibbs sampler retains the empirical performance of full posterior inference while remaining tractable and achieving equivalent regret.
\end{enumerate}

We argue that these contributions are valuable in practice,
as the same bandit algorithm is readily applicable across complex (\eg non-exponential-family) multi-armed bandits,
and attains reduced regret without stringent model design choices or case-by-case hyperparameter tuning, avoiding model misspecification.

Section~\ref{sec:background} reviews Thompson sampling and Bayesian nonparametric mixture models,
before we present in Section~\ref{sec:method} \texttt{Nonparametric TS} and its inference procedure.
Section~\ref{sec:evaluation} describes our empirical study,
organized around the three research questions that support our contributions.
We discuss limitations in Section~\ref{sec:discussion}, and conclude with Section~\ref{sec:conclusion}.
Model extensions, experiments, and reproducibility details are in the Appendix,
with the codebase at \url{https://anonymous.4open.science/r/nonparametric_ts-688D}.

\section{Background}
\label{sec:background}
\paragraph{Contextual multi-armed bandits.}
At each interaction $t$, an agent observes context $x_t \in \mathcal{X}$ and selects an arm $a_t \in \A$, observing reward $y_{t,a_t}$ drawn from an unknown, context-conditional distribution $Y_{t,a_t} \sim p(Y|a_t,x_t,\thetastar)$.
Given the true model,
with parameters $\thetastar$,
the optimal policy maximizes the conditional expected reward $\mu_{t,a}(x_t,\thetastar)=\eValue{p(Y|a,x_t,\thetastar)}{Y}$
across arms.
Equivalently, an agent's goal is to minimize cumulative regret, the expected loss incurred from not knowing the best arm $A_t^*$ at each interaction, $R_T = \eValue{}{\sum_{t=1}^T (Y_{t,A_t^*}-Y_{t,A_t})}$, where the expectation is over the randomness of the outcomes $Y$ and of the (possibly stochastic) arm-selection policy.

\paragraph{Thompson sampling (TS).}
TS~\citep{j-Thompson1933,j-Russo2018} chooses the next arm in proportion to its probability of being optimal, given history $\HH_{1:t-1}=(x_{1:t-1},a_{1:t-1},y_{1:t-1})$.
In its Bayesian formulation, uncertainty about the reward model is captured with a distribution (prior, then posterior) over $\theta$.
TS draws a parameter sample from the posterior given history, $\theta^{(t)} \sim p(\theta|\HH_{1:t-1})$;
and plays the arm that maximizes the resulting expected reward, $a_t = \argmax_{a' \in \A}\mu_{t,a'}(x_t,\theta^{(t)})$.
This is tractable in closed form only when $p(Y|\theta)$ is within the exponential family~\citep{ic-Korda2013,j-Russo2018}.

Two broad themes have pursued extensions of TS to other distributions.
The first aims at explicit generative \emph{reward modeling},
embracing Bayesian neural networks with approximate inference
---variational, mini-batch, Monte Carlo~\citep{ip-Blundell2015,ic-Kingma2015,ip-Lipton2018,ic-Osband2016,ip-Li2016};
an ensemble of plausible models~\citep{ip-Lu2017}, analyzed by~\citet{ic-Qin2022};
and more recently, epistemic neural networks~\citep{ic-Osband2023}.
In all these, posterior reward uncertainty estimation is critical for bandit performance.
The second forgoes explicit density modeling,
instead estimating per-arm reward \emph{means} empirically via optimistic perturbation schemes~\citep{ip-Agrawal2013,ip-Agrawal2013a}, bootstrapping with artificially generated data~\citep{j-Osband2015,j-Eckles2019},
or Dirichlet-weighted re-sampling of each arm's observed (bounded) rewards~\citep{ip-Riou2020},
and its Dirichlet sampling generalization~\citep{dirichletts2021}, all for non-contextual bandits.

Bayesian nonparametrics have separately been used for MABs from two complementary perspectives.
On the one hand, to accommodate continuous or countably infinite \emph{action} spaces, \eg Gaussian processes over a continuum of arms~\citep{ip-Srinivas2010,ip-Gruenewaelder2010,ic-Krause2011} or hierarchical Pitman-Yor priors over an unknown number of arms~\citep{j-Battiston2018}.
On the other hand, \citet{dpps2025} integrate Dirichlet processes within TS
as a prior over the reward distribution's cumulative distribution function (CDF).
This implies almost surely discrete reward distributions,
and requires finite-dimensional approximations to infinite-dimensional CDFs.
In contrast, we model (context-conditional) reward \emph{densities} with DP mixtures of continuous components.

Extending TS to general reward distributions,
while retaining a principled generative account of Bayesian reward uncertainty,
is the focus of this work.

\paragraph{Bayesian nonparametric (BNP) mixture models.}
A BNP model places a prior on an infinite-dimensional parameter space, yet uses only a finite subset of it to explain a finite sample, as the effective number of active dimensions adapts to observed data~\citep{b-Ghosal2017}.
In a Dirichlet process (DP) mixture model~\citep{j-Antoniak1974,j-Escobar1995}, observations $Y_n$ are generated as
\begin{align}
G &\sim DP(\varphi|\gamma,G_0)\;, \; \varphi_n \sim G \;, \;Y_n \sim p(Y|\varphi_n)\;,
\label{eq:dp_generative_background}
\end{align}
where the $DP(\varphi|\gamma,G_0)$ prior~\citep{j-Ferguson1973} is itself a distribution over distributions $G$, with concentration hyperparameter $\gamma>0$ and base measure $G_0$.
A draw $G$ is almost surely discrete, which induces a clustering property among $\varphi_{1:n}$:
repeated values occur with positive probability, so that $n$ draws collapse onto $K \le n$ distinct atoms, with $K$ itself random and growing (logarithmically) with $n$~\citep{j-Antoniak1974,j-Neal2000}.
This is precisely the mechanism that lets a DP mixture model \emph{grow in complexity as needed}: rather than fixing the number of mixture components a priori, %
it is inferred from data.

Crucially, DP Gaussian mixtures are universal and consistent density estimators for continuous distributions:
under mild regularity conditions on the true density and base measure, the DP mixture posterior converges (in total variation, and at rates comparable to minimax-optimal, up to logarithmic factors) to the true, unknown data-generating density as more data are observed~\citep{j-Ghosal1999,j-Ghosal2001,j-Lijoi2004,j-Ghosal2007,j-Bhattacharya2010,j-Ghosal2010},
with related results for conditional densities via covariate dependence~\citep{j-Tokdar2006,j-Pati2013}.
Extensions of the DP such as the Pitman-Yor (PY) process allow for power-law tail behavior~\citep{j-pitmanyor1997,j-Ishwaran2001}.

This established body of BNP theory is the motivation for our approach: it justifies using a BNP Gaussian mixture as a model of an arm's reward distribution, whatever its true (continuous) shape may be,
rather than requiring the bandit designer 
\emph{to correctly guess} a parametric family in advance.

\section{Bayesian Nonparametric Thompson sampling}
\label{sec:method}

We combine Bayesian nonparametric mixture models with Thompson sampling 
for MABs under reward model uncertainty,
by modeling the bandit's unknown per-arm reward distributions with BNP mixture models.
We focus on DP mixtures below;
Appendix~\ref{asec:pitman_yor} records the direct extension to PY mixtures.
For a collection of observed contexts, actions and rewards at time $t$, $\HH_{1:t}=\left(x_{1:t}, a_{1:t}, y_{1:t}\right)$,
we fit a BNP model to the observed data,
and let the TS agent use the BNP posterior predictive distribution $\ptilde(Y|a,x_{t+1},\HH_{1:t})$ to contextually select the next arm.

\subsection{Per-arm rewards as Dirichlet process Gaussian mixtures}
\label{ssec:nonparametric_rewards}

We model each arm's context-conditional reward distribution with an independent DP Gaussian mixture,
\begin{align}
G_{a} &\sim DP(\varphi_a|\gamma_a, G_{a,0}) \;, \label{eq:per_arm_dp} \\
\varphi_{t,a} &\sim G_{a} \;, \label{eq:per_arm_dp_param}\\
Y_{t,a} &\sim p(Y|x_t,\varphi_{t,a})=\N{Y|x_t^\top w_{t,a}, \sigma_{t,a}^2} \;, \label{eq:per_arm_dp_observations}
\end{align}
with $\varphi_{t,a}=(w_{t,a},\sigma_{t,a}^2)$ and a normal inverse-gamma base measure $G_{a,0}(\varphi_a|\varPhi_{a,0}) = \N{w_a|U_{a,0},\sigma_a^2 V_{a,0}}\IG{\sigma_a^2|\alpha_{a,0},\beta_{a,0}}$, conjugate to the Gaussian emission, which enables closed-form posterior updates (Appendix~\ref{asec:nonparametric_ts_details}).
Figure~\ref{fig:pgm_nonparametric_bandit} shows the BNP graphical model: each arm's reward distribution $G_a$ is estimated \emph{independently},
so that different arms may follow entirely different reward distributions.\footnote{
	The alternative \emph{hierarchical} BNP model described in Appendix~\ref{asec:hierarchical} shares a base measure across arms.
}

\begin{figure}[!h]
\vspace{-2ex}
\begin{center}
\scalebox{0.9}{\begin{tikzpicture}
	\node[obs] (y-t) {$y_{t}$};
	\node[latent, above=0.6 of y-t, xshift=0cm] (theta-ta) {$\varphi_{t,a}$};
	\node[latent, above=0.5 of theta-ta, xshift=0cm] (G-a) {$G_{a}$};
	\node[latent, right=1 of y-t] (a-t) {$a_t$};
	\node[latent, left=1 of y-t] (x-t) {$x_t$};
	
	\node[const, above=0.1 of G-a, xshift=-1.0cm] (gamma-a) {$\gamma_{a}$} ;
	\node[const, above=0.1 of G-a, xshift=1.0cm]  (G-a0) {$G_{a,0}$} ;
	
	\edge {gamma-a,G-a0} {G-a} ;
	\edge {G-a} {theta-ta} ;
	\edge {theta-ta,x-t,a-t} {y-t} ;
	
	\plate {t} {(theta-ta)(a-t)(x-t)(y-t)} {$t$} ;
	\plate {a}{
		(gamma-a)(G-a0) %
		(G-a) %
		(theta-ta) %
	} {$A$} ;
\end{tikzpicture}}
\caption{Graphical model of the per-arm DP Gaussian mixture bandit.}
\label{fig:pgm_nonparametric_bandit}
\end{center}
\vspace{-3ex}
\end{figure}

Equation~\eqref{eq:per_arm_dp_observations} is a DP mixture of linear regressions that flexibly models the shape of each arm's conditional reward density (\eg multimodal, skewed or outlier-contaminated) around a conditional mean that is linear in the context.
We illustrate BNP modeling flexibility in the standard contextual linear setting (leaving nonlinear mean functions as future work);
for non-contextual bandits, setting $x_t=1$ reduces the model to an unconditional DP Gaussian mixture density estimator (see further details in Appendix~\ref{assec:nonparametric_ts_expressivity}).

\paragraph{Predictive posterior.}
Let $(t_a)=(\tau\leq t: a_\tau=a)$ denote the sequence of time-instants
where arm $a$ has been played up to $t$,
for a total of $n_{t,a}$ instances,
and $y_{(t_a)}$ the corresponding observed rewards.
Marginalizing $G_a$ yields the per-arm predictive reward distribution\footnote{
	The predictive posterior generalizes directly to Pitman-Yor priors, with power-law tails (Appendix~\ref{asec:pitman_yor}).
},
conditioned on parameters $\varphi_{t,a}$, the current mixture assignments of past rewards, and context $x_{t+1}$:
\begin{align}
&Y_{t+1,a} | x_{t+1}, y_{(t_a)}, \varphi_{t,a}, \gamma_a, G_{a,0} \sim \nonumber\\
&\qquad \sum_{k=1}^{K_{t,a}} \frac{n_{t,a,k}}{n_{t,a}+\gamma_a} \N{Y|x_{t+1}^\top w_{a,k}^*, \sigma_{a,k}^{*2}} \nonumber \\
&\qquad + \frac{\gamma_a}{n_{t,a}+\gamma_a} \N{Y|x_{t+1}^\top w_{a,k_{new}}, \sigma_{a,k_{new}}^2} \;,
\label{eq:nonparametric_Gaussian_mixture_detailed}
\end{align}
where $\varphi_{a,k}^*=(w_{a,k}^*,\sigma_{a,k}^{*2})$, $k=1,\dots,K_{t,a}$, are the \textit{posterior parameters} of the $K_{t,a}$ mixture components already instantiated by the observed data.
Each $\varphi_{a,k}^*$ follows a normal inverse-gamma posterior,
conjugate-updated from $\varPhi_{a,0}$ (see Appendix~\ref{assec:nonparametric_ts_posterior}).
$\varphi_{a,k_{new}}=(w_{a,k_{new}}, \sigma_{a,k_{new}}^2) \sim G_{a,0}$ is a candidate new component drawn from the prior base measure per-arm.
The number of active components $K_{t,a}$ is not fixed, but grows as needed to explain each arm's observed rewards $y_{(t_a)}$:
the mechanism by which the model sidesteps having to specify a reward family in advance.

\subsection{Nonparametric Thompson sampling}
\label{ssec:nonparametric_thompson_sampling}

At each bandit interaction,
\texttt{Nonparametric TS} as in Algorithm~\ref{alg:nonparametric_ts}
draws a parameter sample from each per-arm BNP posterior.
Because the marginalized predictive distribution mixes over both instantiated
($k=1,\dots,K_{t,a}$) and new ($k_{new}$) components,
posterior parameters\footnote{
These posteriors can be computed sequentially as data are observed for the played
arm.}
cover both cases, \ie
\begin{align}
	\varphitilde_a^{(t)} &:
	\begin{cases}
		\varphitilde_{a,k}^{*(t)} \sim G_{a,n_{t,a,k}}(\varphi_{a,k}^*|\varPhi_{a,k,n_{t,a,k}}), \\
		\varphitilde_{a,k_{new}}^{(t)} \sim G_{a,0}(\varphi_a|\varPhi_{a,0})\;.
	\end{cases}
	\label{eq:nonparametric_posterior_draw}
\end{align}

\texttt{Nonparametric TS} then computes per-arm expected rewards under the predictive distribution of Equation~\eqref{eq:nonparametric_Gaussian_mixture_detailed},
\ie
\begin{align}
\mu_{t,a}(x_t,\varphitilde_a^{(t)}) &= \sum_{k=1}^{K_{t,a}} \frac{n_{t,a,k}}{n_{t,a}+\gamma_a}\left(x_t^\top \tilde{w}_{a,k}^{(t)*}\right) \nonumber \\
&\quad + \frac{\gamma_a}{n_{t,a}+\gamma_a}\left(x_t^\top \tilde{w}_{a,k_{new}}^{(t)}\right) \;,
\label{eq:dp_expected_reward}
\end{align}
and plays the arm with highest sampled expected reward
$a_t = \argmax_{a'\in\A}\mu_{t,a'}(x_t,\varphitilde_{a'}^{(t)})$.

\begin{algorithm}[!h]
\caption{Nonparametric Thompson sampling}
\label{alg:nonparametric_ts}
\begin{algorithmic}[1]
\STATE {\bfseries Input:} $\A$; per-arm BNP hyperparameters $\gamma_a,\varPhi_{a,0}$; Gibbs budget $Gibbs_{max}$ and $\epsilon$.
\STATE $\HH_1=\emptyset$
\FOR{$t=1,\dots,T$}
\STATE Receive context $x_t$
\FOR{$a \in \A$}
\STATE Draw $\varphitilde_a^{(t)}$ from the posterior in Eq.~\eqref{eq:nonparametric_posterior_draw}
\STATE Compute $\mu_{t,a}(x_t,\varphitilde_a^{(t)})$ via Eq.~\eqref{eq:dp_expected_reward}
\ENDFOR
\STATE Play $a_t = \argmax_{a'\in\A}\mu_{t,a'}(x_t,\varphitilde_{a'}^{(t)})$
\STATE Observe $y_t$, update $\HH_{1:t} = \HH_{1:t-1}\cup\{x_t,a_t,y_t\}$
\STATE Update arm $a_t$'s posterior with Algorithm~\ref{alg:nonparametric_gibbs}
\ENDFOR
\end{algorithmic}
\end{algorithm}

\subsection{BNP posterior inference: a warm-started, budgeted Gibbs sampler}
\label{ssec:inference}

Computing Equations~\eqref{eq:nonparametric_Gaussian_mixture_detailed}, \eqref{eq:nonparametric_posterior_draw}, \eqref{eq:dp_expected_reward}
requires inferring latent mixture assignments $z_{1:t}$,
to compute how many observations $n_{t,a,k}$ per-arm $a$ at time $t$ are assigned to component $k$.
We use a marginalized Gibbs sampler
(pseudo-code in Algorithm~\ref{alg:nonparametric_gibbs},
full procedure details in Appendix~\ref{assec:nonparametric_ts_inference})
following~\citet{j-Neal2000}, which alternates between resampling each observation's assignment and updating the affected component's conjugate posterior, until a relative log-likelihood change below $\epsilon$ is reached or a maximum of $Gibbs_{max}$ iterations is used.

\begin{algorithm}[!h]
\caption{Gibbs sampler for arm $a$'s posterior}
\label{alg:nonparametric_gibbs}
\begin{algorithmic}[1]
\STATE {\bfseries Input:} BNP prior hyperparameters $\gamma_a,\varPhi_{a,0}$;
observations $(x_{(t_a)},y_{(t_a)})$;
$\epsilon,Gibbs_{max}$
\WHILE{not converged \& iterations $< Gibbs_{max}$}
\FOR{$t' \in (t_a)$}
\STATE Resample $z_{t'} \in \{1,\dots,K_{t,a},k_{new}\}$ $\propto$ (mixture weights $\times$ likelihood of $y_{t'}$ per-k)
\STATE Update sufficient statistics $n_{t,a,k}$
\STATE Compute posterior parameters $\varphi_{a,k}^*$ of affected component(s) $k$.
\ENDFOR
\STATE Recompute joint log-likelihood
\ENDWHILE
\end{algorithmic}
\end{algorithm}

Two design choices keep this BNP inference practical,
which we validate empirically in Section~\ref{ssec:rq3}.
First, because only one arm is played per MAB interaction,
the sampler is \emph{warm-started} at every interaction from the assignments and posterior parameters
already inferred for that arm's previous $n_{t,a}-1$ rewards,
avoiding re-initialization from the prior.
Second, we cap the number of Gibbs iterations at $Gibbs_{max}$ per MAB-interaction:
each interaction then costs at most $\mathcal{O}(Gibbs_{max}\, n_{t,a}\,(K_{t,a}+1))$ density evaluations.
Since the expected number of components under the DP prior grows logarithmically with the number of observations, $\mathbb{E}[K_{t,a}]\approx\gamma_a\log(1+n_{t,a}/\gamma_a)$~\citep{j-Antoniak1974},
the expected per-interaction a priori cost is $\mathcal{O}(Gibbs_{max}\, n_{t,a}\log n_{t,a})$.

\section{Empirical Study}
\label{sec:evaluation}

We evaluate \texttt{Nonparametric TS}
and compare it to alternatives described below,
in a variety of previously elusive scenarios
spanning multimodal, outlier-contaminated, and non-exponential-family reward distributions;
including bandits with \emph{different} reward distributions per arm
(see Table~\ref{tab:evaluation_scenarios}).
We organize our empirical study around three research questions.
Section~\ref{ssec:rq1} inquires whether
\texttt{Nonparametric TS} recovers Oracle-level performance,
despite not knowing the true reward model, and what a misspecified reward model costs.
Section~\ref{ssec:rq2} evaluates how well \texttt{Nonparametric TS} compares to state-of-the-art baselines under reward model uncertainty.
Finally, 
Section~\ref{ssec:rq3} investigates whether \texttt{Nonparametric TS}'s BNP flexibility remains computationally practical.

\begin{table*}[t]
	\caption{Evaluated scenarios and corresponding Oracle TS.}
	\label{tab:evaluation_scenarios}
	\centering
	\vspace*{-2ex}
	\small
	\begin{tabular}{lcc}
		\toprule
		Scenario & Reward model & Oracle TS \\
		\midrule
		Contextual linear Gaussian & Eq.~\eqref{eq:per_arm_dp_observations} & \citet{ip-Agrawal2013a} \\
		Gaussian mixture (\texttt{A}, \texttt{B}) & Eq.~\eqref{eq:scenario_mixture_A}--\eqref{eq:scenario_mixture_B} & \citet{ip-Urteaga2018}$^\dagger$\\
		Outlier-contaminated & Eq.~\eqref{eq:scenario_heavy_tail} & \citet{ip-Urteaga2018}$^\dagger$ \\
		Exponentially distributed & Eq.~\eqref{eq:scenario_exponential} & N/A \\
		\bottomrule
	\end{tabular}
	\\[0.5ex]
	{\footnotesize $^\dagger$Mixture model of~\citet{ip-Urteaga2018} with known $K_a$, inferred with the same Gibbs sampler as in Section~\ref{ssec:inference}.}
\end{table*}

\subsection{Setup}
\label{ssec:evaluation_setup}

We evaluate \texttt{Nonparametric TS} with a DP Gaussian mixture per arm
($\gamma_a=0.1$, $Gibbs_{max}=10$, $\epsilon=10^{-4}$)
---sensitivity to $\gamma_a$ and $Gibbs_{max}$ studied in Appendices~\ref{asec:gamma} and~\ref{asec:gibbs}---
and compare it against 9 baselines evaluated by~\citet{ip-Riquelme2018},
spanning conjugate linear-Gaussian TS~\citep{ip-Agrawal2013a}, multi-task Gaussian processes~\citep{ip-Srinivas2010,ip-Gruenewaelder2010,ic-Krause2011}, and seven neural-network based approaches: a learned-representation linear TS, bootstrapping, RMSProp and dropout-based perturbation, parameter noise injection, and two Bayesian neural networks with variational and expectation-propagation inference.

All baselines are implemented and set with default hyperparameters as reported by~\citet{ip-Riquelme2018},
with full descriptions in Appendix~\ref{asec:baselines}.
We report cumulative regret $R_t$ as defined in Section~\ref{sec:background} over $N\!\geq\!500$ independent realizations:
figures show its mean (standard deviation shaded) over time,
and tables its mean $\pm$ standard deviation at the final interaction $T$.

\subsection{RQ1: Does \texttt{Nonparametric TS} recover Oracle performance?}
\label{ssec:rq1}

MAB model uncertainty is the premise of this work:
a bandit designer rarely knows the true per-arm reward distribution in advance, and committing to the wrong parametric family costs regret.
An Oracle TS that is handed the true model class is therefore an \textit{unrealistic} benchmark of achievable performance.
Yet we evaluate below how close \texttt{Nonparametric TS}'s regret is to per-scenario Oracle TS implementations (see Table~\ref{tab:evaluation_scenarios}),
to assess 
whether the BNP posterior is successful in what we use it for,
\ie estimating unknown densities.

We first verify this with no model mismatch.
We start with 12 parameterizations of contextual linear Gaussian bandits
for $|\A|\in \{2,\dots,5\}$ with varying arm separation.
We observe that \texttt{Nonparametric TS} matches the conjugate,
correctly-specified \texttt{LinearGaussian TS} in both mean regret and volatility.
Regret curves are visually indistinguishable throughout all panels in Appendix~\ref{asec:linear_oracle},
and neither policy dominates the other:
final mean cumulative regrets are within $\approx 3\%$ of each other across all scenarios,
a gap dwarfed by the standard deviation bands.

We positively conclude that the BNP predictive posterior loses little when the truth is \emph{simple} (linear-Gaussian).
We also test the more interesting case when the truth is \emph{not simple},
\eg not in the exponential family.
We simulate two contextual Gaussian-mixture scenarios,
with %
context $x_{t}\sim\U{0,1}^2$,
and reward densities $p_a(y|x_t,\theta)$ for arm $a$
(denoted as $p_a$):
{\footnotesize
\begin{align}
&\texttt{A:}
\begin{cases}
p_1 = 0.5\N{y|[1\;1]^\top x_t,1}+0.5\N{y|[2\;2]^\top x_t,1} \\
p_2 = 0.3\N{y|[0\;0]^\top x_t,1}+0.7\N{y|[3\;3]^\top x_t,1}
\end{cases} \hspace*{-2ex}
\label{eq:scenario_mixture_A}
\end{align}
\vspace*{-2ex}
\begin{align}
&\texttt{B:}
\begin{cases}
p_1 = \N{y|[1\;1]^\top x_t,1} \\
p_2 = 0.5\N{y|[1\;1]^\top x_t,1}+0.5\N{y|[2\;2]^\top x_t,1} \\
p_3 = 0.3\N{y|[0\;0]^\top x_t,1}+0.6\N{y|[3\;3]^\top x_t,1} \\
\hspace{0.34\linewidth} + 0.1\N{y|[4\;4]^\top x_t,1}
\end{cases} \hspace*{-2ex}
\label{eq:scenario_mixture_B}
\end{align}
}

\begin{figure}[!h]
\centering
\includegraphics[width=0.48\linewidth]{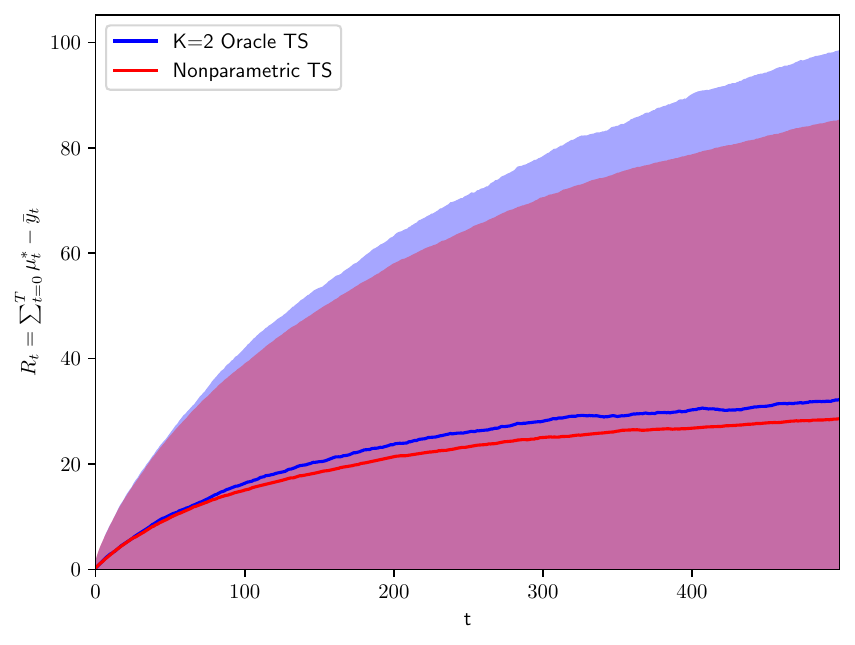}
\includegraphics[width=0.48\linewidth]{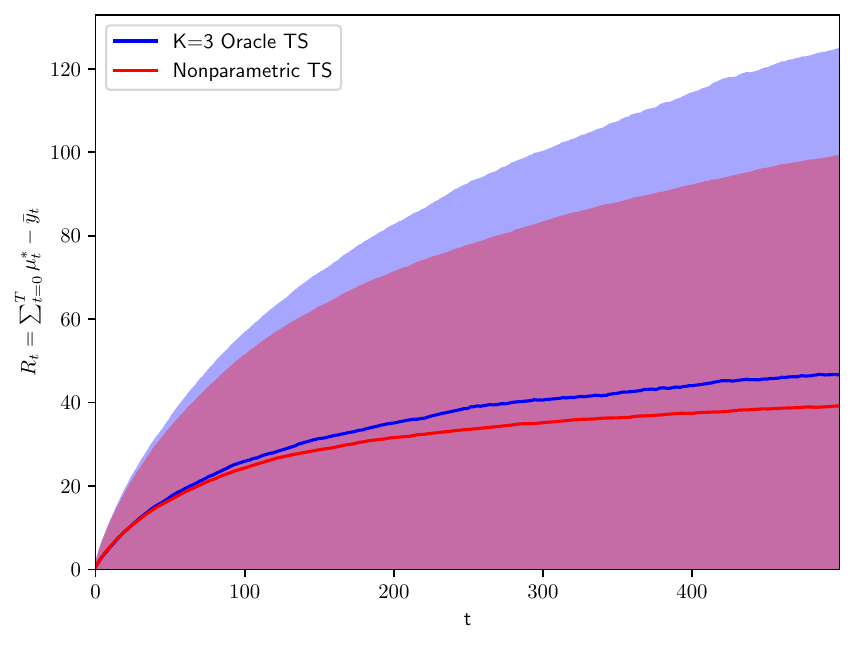}
\vspace{-1ex}
\caption{Mean $R_t$ (std.\ shaded) of \texttt{Nonparametric TS} vs.\ Oracle TS, Scenarios \texttt{A} (left) and \texttt{B} (right).}
\label{fig:mixture_scenarios_oracle}
\end{figure}

\texttt{Scenario A} has two arms with significantly overlapping rewards, one with unbalanced mixture weights;
while \texttt{Scenario B} covers \emph{different} distributions per arm 
---unimodal, bimodal, and unbalanced three-component mixtures, respectively.
Figure~\ref{fig:mixture_scenarios_oracle} compares \texttt{Nonparametric TS} to per-scenario Oracle TS policies that know the true number of mixture components $K_a$ in each case.
We observe that \texttt{Nonparametric TS} attains regret comparable to the Oracle in both scenarios,
despite using the \emph{same} algorithm and hyperparameters in both scenarios, whereas the Oracle must be configured per scenario.

Finally, we assess the cost of \emph{model misspecification},
on a bandit whose per-arm rewards are a two-component mixture with a shared mean, one component having inflated variance,
producing outlier-contaminated rewards:
{\footnotesize
\begin{align}
\hspace*{-2ex}
\begin{cases}
p_1 = 0.75\N{y|[0\;0]^\top x_t,1}+0.25\N{y|[0\;0]^\top x_t,100} \\
p_2 = 0.75\N{y|[2\;2]^\top x_t,1}+0.25\N{y|[2\;2]^\top x_t,100}
\end{cases} \hspace*{-3ex}
\label{eq:scenario_heavy_tail}
\end{align}
}

\begin{figure}[!h]
\centering
\includegraphics[width=0.48\linewidth]{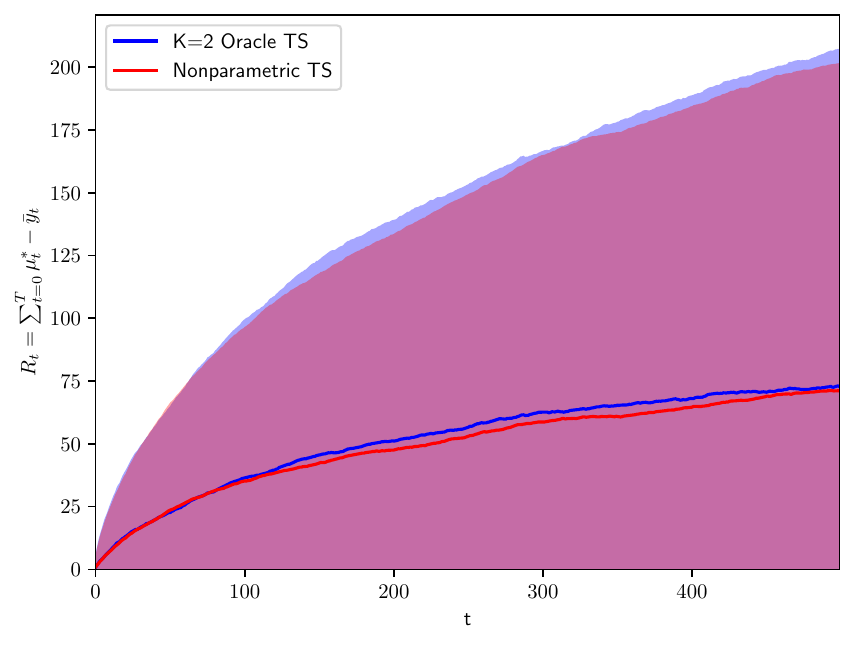}
\includegraphics[width=0.48\linewidth]{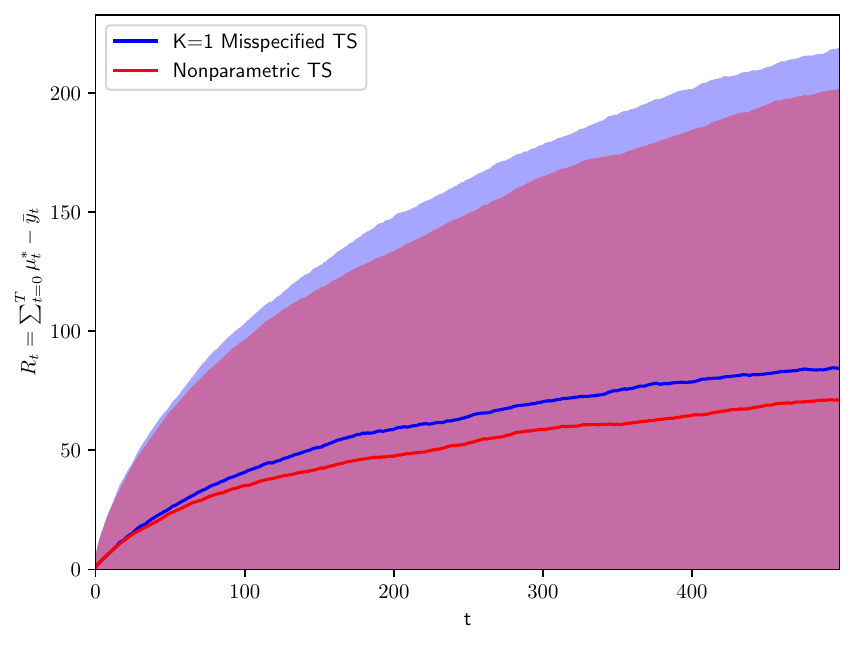}
\vspace{-1ex}
\caption{Outlier-contaminated bandit: Mean $R_t$ (std.\ shaded) of \texttt{Nonparametric TS} vs.\ a correctly-specified Oracle (left), vs.\ a misspecified Gaussian TS (right).}
\label{fig:heavy_tail}
\end{figure}

Figure~\ref{fig:heavy_tail} compares, on this bandit,
an Oracle that correctly knows $K_a=2$ mixture components per arm,
against a misspecified TS that assumes a unimodal Gaussian.
The misspecified policy incurs 18\% higher cumulative regret than \texttt{Nonparametric TS}, which needs no such prior knowledge and matches the Oracle throughout.
This evidence is the clearest illustration of what reward-model uncertainty costs,
and what modeling it via BNP mixture models provides.

\subsection{RQ2: How does \texttt{Nonparametric TS} compare to baselines?}
\label{ssec:rq2}

For contextual linear Gaussian bandits, a correctly-specified conjugate baseline exists, \ie \texttt{LinearGaussian TS} is the Oracle TS.

This scenario tests whether \emph{flexible} TS algorithms pay 
a price for their generality.
Table~\ref{tab:linear_showdown_baselines_regret}
($|\A|=8$, $x_t\in\Real^{10}$, dense $\theta_a$; sparse $\theta_a$ and figures in Appendix~\ref{asec:linear_baselines}) shows that the baselines do: \texttt{NeuralLinear} incurs roughly double the cumulative regret of \texttt{Nonparametric TS}, and the two Bayesian neural network baselines more than 300\% additional regret.

\begin{table}[!h]
\caption{$R_T$ at $T=1500$, contextual linear Gaussian bandit (dense $\theta_a$). Parentheses: relative regret increase over \texttt{Nonparametric TS}. Selected baselines; full 9-baseline table in Appendix~\ref{asec:linear_baselines}.}
\label{tab:linear_showdown_baselines_regret}
\vspace{-2ex}
\centering
\resizebox{\linewidth}{!}{
\begin{tabular}{lr}
\toprule
Algorithm & Cumulative regret (rel.) \\
\midrule
\textbf{Nonparametric TS} & \textbf{189.9 $\pm$ 13.0} (0\%) \\
NeuralBootstrapped & 210.9 $\pm$ 47.4 (11\%) \\
MultitaskGP & 213.6 $\pm$ 22.0 (13\%) \\
NeuralLinear & 398.5 $\pm$ 18.2 (110\%) \\
BNNVariationalGaussian & 827.4 $\pm$ 193.1 (336\%) \\
\bottomrule
\end{tabular}}
\end{table}
\vspace{-1ex}

\begin{figure}[!h]
\centering
\includegraphics[width=0.48\linewidth]{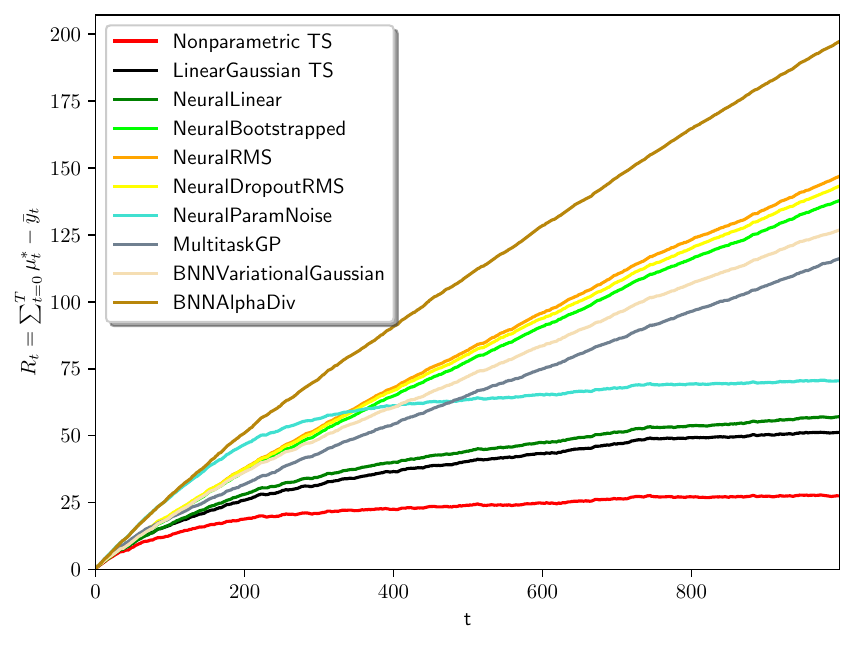}
\includegraphics[width=0.48\linewidth]{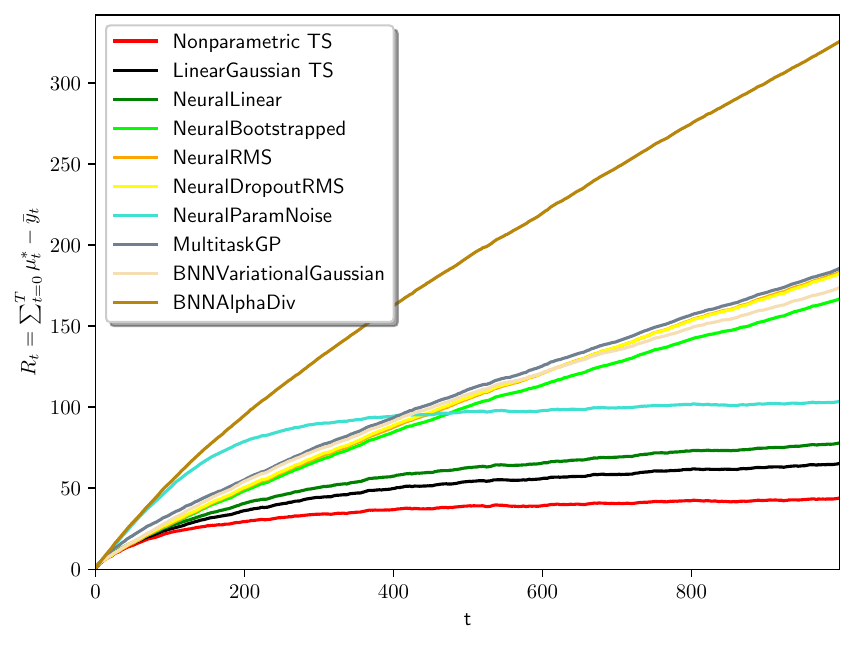}
\vspace{-1ex}
\caption{Mean $R_t$, Scenarios \texttt{A} (left) and \texttt{B} (right).}
\label{fig:mixture_scenarios_bandit_showdown_baselines_regret}
\end{figure}

The gap to most baselines widens with increased reward-model complexity:
Figure~\ref{fig:mixture_scenarios_bandit_showdown_baselines_regret} and Table~\ref{tab:mixture_scenarios_bandit_showdown_baselines_regret} show cumulative regret in scenarios \texttt{A} and \texttt{B} of Eq.~\eqref{eq:scenario_mixture_A} and \eqref{eq:scenario_mixture_B}.
Only \texttt{Nonparametric TS}, \texttt{LinearGaussian TS}, \texttt{NeuralLinear} and \texttt{NeuralParamNoise}
reach a plateauing (\ie converged) regret curve.
\texttt{Nonparametric TS} converges fastest and to the lowest regret,
with \texttt{NeuralLinear} incurring 77--108\% more regret.
Bootstrapped, RMS and BNN-based baselines fail to find the exploration-exploitation balance, incurring 4 to 7$\times$ higher cumulative regret with markedly larger volatility.

\begin{table}[!h]
\caption{$R_T$ at $T=1000$, mixture scenarios. Parentheses: relative regret increase over \texttt{Nonparametric TS}. Selected baselines; full table in Appendix~\ref{asec:mixture_baselines}.}
\label{tab:mixture_scenarios_bandit_showdown_baselines_regret}
\vspace{-2ex}
\centering
\resizebox{\linewidth}{!}{
\begin{tabular}{lrr}
\toprule
Algorithm & Scenario \texttt{A} & Scenario \texttt{B} \\
\midrule
\textbf{Nonparametric TS} & \textbf{27.5 $\pm$ 70.7} (0\%) & \textbf{44.1 $\pm$ 70.4} (0\%) \\
LinearGaussian TS & 51.2 (87\%) & 65.4 (48\%) \\
NeuralLinear & 57.0 (108\%) & 78.1 (77\%) \\
BNNAlphaDiv & 197.4 (619\%) & 325.7 (638\%) \\
\bottomrule
\end{tabular}}
\end{table}
\vspace{-1ex}

\begin{figure}[!h]
\centering
\includegraphics[width=0.9\linewidth]{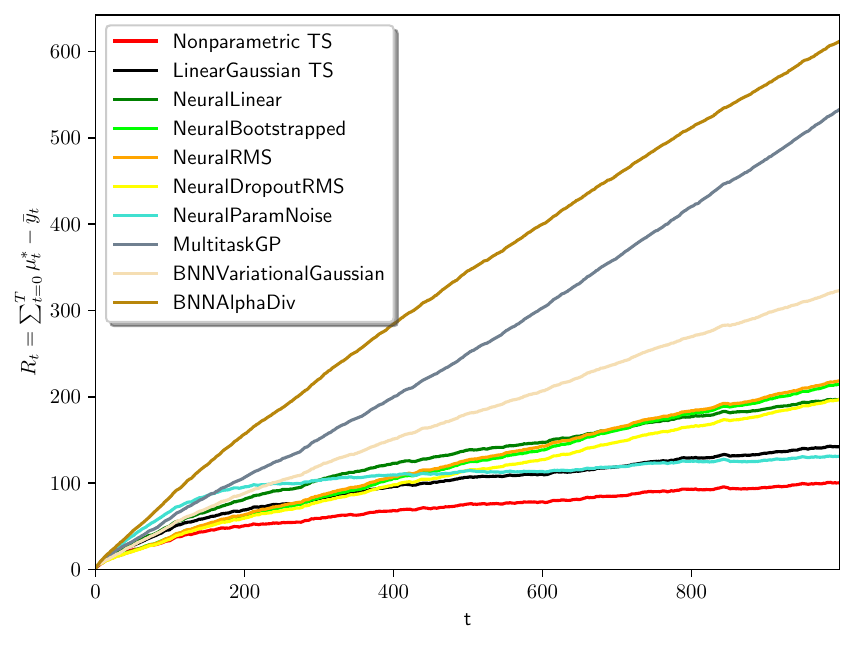}
\vspace{-1ex}
\caption{Mean $R_t$, outlier-contaminated bandit.}
\label{fig:heavy_tailed_regret}
\end{figure}

\paragraph{Outlier-contaminated rewards.}
With outlier-contaminated rewards as in Eq.~\eqref{eq:scenario_heavy_tail}, only \texttt{Nonparametric TS}, \texttt{NeuralParamNoise} and \texttt{LinearGaussian TS} find the right exploration-exploitation balance:
see Figure~\ref{fig:heavy_tailed_regret} and Table~\ref{tab:heavy_tailed_baselines_regret}.

\texttt{Nonparametric TS} does so fastest,
with \texttt{NeuralParamNoise} and \texttt{LinearGaussian TS} incurring 31\% and 42\% more cumulative regret, respectively,
while the remaining baselines incur roughly twice its regret or more,
most of them with markedly larger volatility.

\begin{table}[!h]
\caption{$R_T$ at $T=1000$, outlier-contaminated bandit. Selected baselines; full table in Appendix~\ref{asec:heavy_tail_baselines}.}
\label{tab:heavy_tailed_baselines_regret}
\vspace{-2ex}
\centering
\resizebox{\linewidth}{!}{
\begin{tabular}{lr}
\toprule
Algorithm & Cumulative regret (rel.) \\
\midrule
\textbf{Nonparametric TS} & \textbf{100.0 $\pm$ 188.9} (0\%) \\
NeuralParamNoise & 130.7 (31\%) \\
LinearGaussian TS & 142.1 (42\%) \\
MultitaskGP & 532.7 (433\%) \\
BNNAlphaDiv & 611.6 (512\%) \\
\bottomrule
\end{tabular}}
\end{table}
\vspace{-1ex}

\begin{figure}[!h]
\centering
\includegraphics[width=0.9\linewidth]{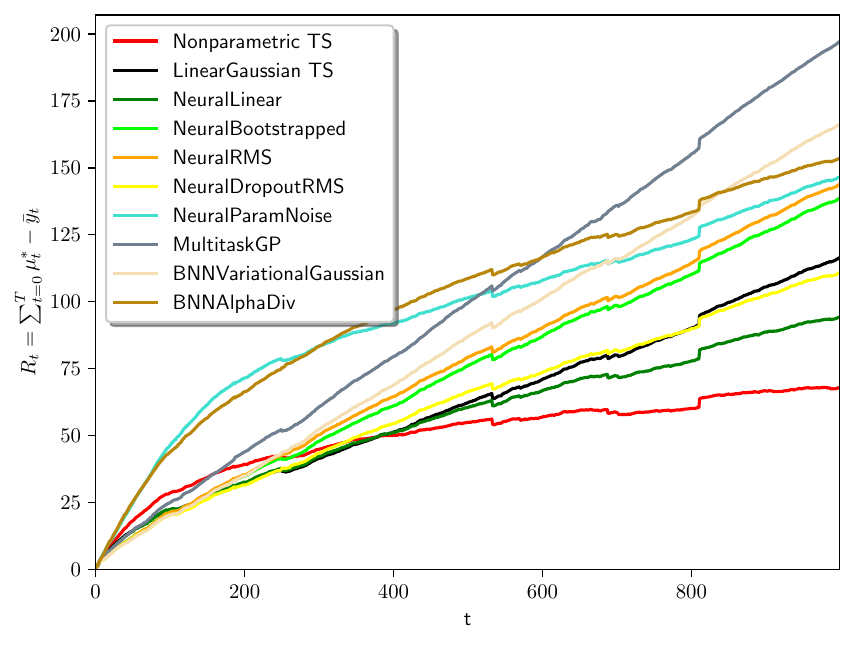}
\vspace{-1ex}
\caption{Mean $R_t$, exponential-reward bandit.}
\label{fig:exponential}
\end{figure}

\paragraph{Exponentially distributed rewards.}
Finally, we test rewards outside the Gaussian family,
by simulating exponentially distributed positive rewards $y\geq0$,
with 2-dimensional context-dependent rates per arm:
\begin{align}
p_a(y|x_t,\theta) = \lambda_a e^{-y\lambda_a}\;, \; \lambda_a = \theta_a^\top x_t\,, \; a=1,2,3\,.
\label{eq:scenario_exponential}
\end{align}
These rewards are contextual in both their mean ($\lambda_a^{-1}$) and variance ($\lambda_a^{-2}$),
with $x_t \sim \U{0,1}^2$ and $\theta_a = a \cdot [1,1]^\top$.
Because no Oracle TS is available for this scenario,
these results speak to both the strength of the challenge and the raw competitiveness of TS baselines:
\texttt{Nonparametric TS} is the only policy that consistently finds the optimal arm (Figure~\ref{fig:exponential} and Appendix~\ref{asec:exponential_baselines}), and every baseline incurs at least 39\% more cumulative regret, up to 191\% for \texttt{MultitaskGP} (Table~\ref{tab:exponential_regret}).

\begin{table}[!h]
\caption{$R_T$ at $T=1000$, exponential-reward bandit. Selected baselines; full table in Appendix~\ref{asec:exponential_baselines}.}
\label{tab:exponential_regret}
\vspace{-2ex}
\centering
\resizebox{\linewidth}{!}{
\begin{tabular}{lr}
\toprule
Algorithm & Cumulative regret (rel.) \\
\midrule
\textbf{Nonparametric TS} & \textbf{67.9 $\pm$ 119.4} (0\%) \\
NeuralLinear & 94.2 (39\%) \\
LinearGaussian TS & 116.4 (71\%) \\
BNNAlphaDiv & 153.5 (126\%) \\
MultitaskGP & 197.2 (191\%) \\
\bottomrule
\end{tabular}}
\end{table}

\subsection{RQ3: Is \texttt{Nonparametric TS} computationally practical?}
\label{ssec:rq3}

A natural concern with Bayesian nonparametric inference is computational cost: unlike the conjugate \texttt{LinearGaussian TS} baseline, \texttt{Nonparametric TS} requires Gibbs sampling at every interaction.
We show it is \emph{tractable}, with a cost directly controllable via the warm-started sampler's budget $Gibbs_{max}$ (Section~\ref{ssec:inference}), without sacrificing the results of Sections~\ref{ssec:rq1}--\ref{ssec:rq2};
we do not claim it is the fastest method (\eg conjugate closed-form TS is, when available).

\begin{figure}[!h]
	\centering
	\includegraphics[width=0.48\linewidth]{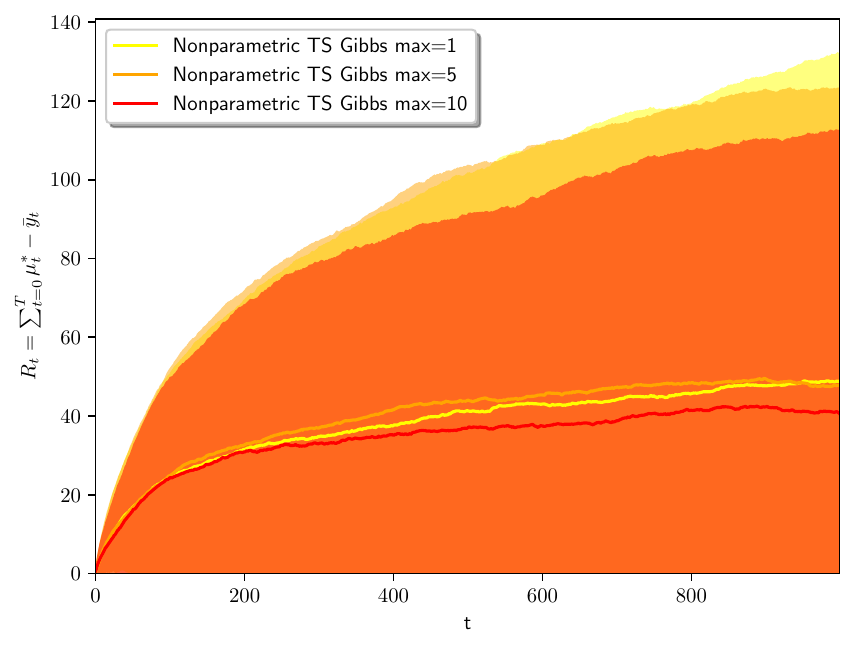}
	\includegraphics[width=0.48\linewidth]{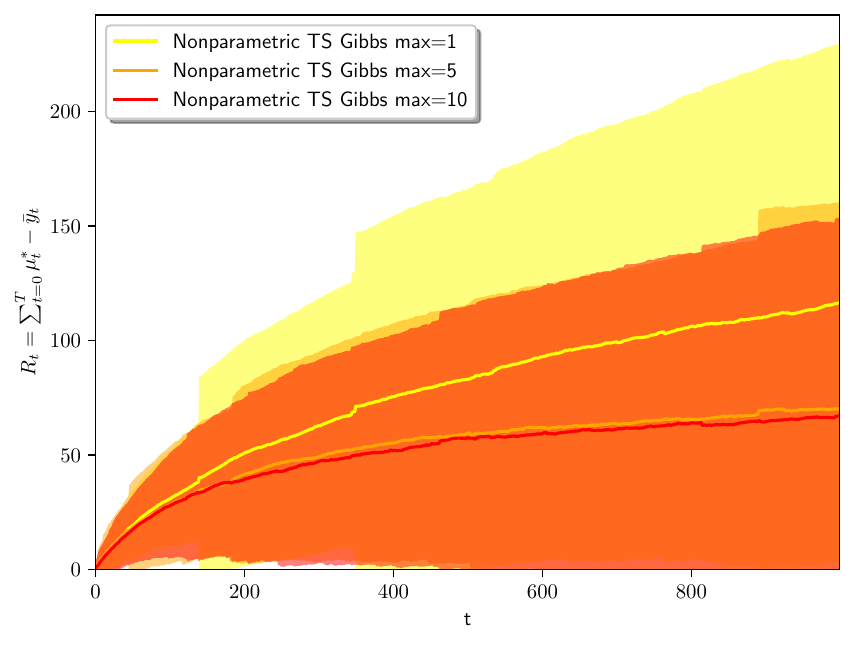}
	\vspace{-1ex}
	\caption{Mean $R_t$ (std.\ shaded) of \texttt{Nonparametric TS} with $Gibbs_{max}\in\{1,5,10\}$, mixture Scenario \texttt{B} (left) and exponential rewards (right).}
	\label{fig:gibbsmaxiter_main}
\end{figure}

Figure~\ref{fig:gibbsmaxiter_main} shows that this cap costs little in regret: $Gibbs_{max}=5$ and $Gibbs_{max}=10$ attain comparable regret
across scenarios of Sections~\ref{ssec:rq1}--\ref{ssec:rq2} (see more results in Appendix~\ref{asec:gibbs}).
Only $Gibbs_{max}=1$ ---\ie sequentially assigning each new reward, without ever revisiting past assignments--- measurably hurts, most clearly for exponentially distributed rewards.
As reported in Appendix~\ref{asec:runtime}, a smaller $Gibbs_{max}$ substantially reduces wall-clock run-time, \eg $Gibbs_{max}=5$ is about 2.5$\times$ faster than $Gibbs_{max}=10$ in the mixture scenarios.
In practice, we recommend $Gibbs_{max} \in [5,10]$ as a good default that keeps inference cheap with satisfactory empirical results, as above.

We separately verify (Appendix~\ref{asec:gamma}) that performance shows only modest sensitivity to the DP concentration hyperparameter $\gamma_a\in[0.01,10]$, so that a single setting $\gamma_a=0.1$ is used across all scenarios.

\section{Discussion}
\label{sec:discussion}
Across a variety of challenging MAB scenarios described in Section~\ref{sec:evaluation},
\texttt{Nonparametric TS} performs on par with each corresponding Oracle TS.
It is not only competitive on average regret, but consistently the \emph{least} (or among the least) volatile among state-of-the-art baselines:
across our simulated scenarios, the standard deviations of cumulative regret of the bootstrapped, RMS, dropout and variational BNN baselines are 1.5--16$\times$ larger,
and only \texttt{BNNAlphaDiv} attains comparable volatility, at 6--7$\times$ higher average regret.

For real-world deployment this matters as much as average regret: a policy whose worst-case realizations are far from its average is harder to trust.

We argue that these regret and volatility advantages, together with the Oracle-equivalence results of Section~\ref{ssec:rq1}, are consistent with our motivating hypothesis (Section~\ref{sec:background}):
a BNP posterior converging to the true reward density should behave satisfactorily across realizations, better than approximating reward densities or estimating means heuristically.

\paragraph{Limitations.}
Our baseline suite follows the implementations and default hyperparameters of~\citet{ip-Riquelme2018} rather than being tuned per scenario.
This keeps the comparison uniform,
but hyperparameter optimization will likely improve neural-network based baselines, so the size of the reported gaps should be read with that in mind.
We emphasize that \texttt{Nonparametric TS} performs well throughout using the same BNP hyperparameters across all scenarios,
with ablations in Appendices~\ref{asec:gamma} and~\ref{asec:gibbs} showing modest sensitivity to $\gamma_a$ and the Gibbs budget.

No Oracle TS is available for exponentially distributed rewards, so that evidence is comparative rather than Oracle-relative.
Our real-world precision oncology evaluation (Appendix~\ref{asec:oncology}) is a considerably harder stress test without ground-truth,
a reminder that our simulated scenarios, while diverse, do not exhaust real-world reward complexity.

We make no formal regret guarantees;
however, we view the Oracle-equivalence results of Section~\ref{ssec:rq1} as the empirical counterpart of such a guarantee.
Establishing a rigorous regret bound for this class of methods is an important, separate, research direction.

\section{Conclusions}
\label{sec:conclusion}
We presented \texttt{Nonparametric TS}, a Thompson sampling algorithm that models each bandit arm's reward distribution with an independent Bayesian nonparametric (\eg Dirichlet process) Gaussian mixture model.
We empirically tested whether this BNP-based reward model, motivated by DP Gaussian mixtures being universal and consistent density estimators, translates into satisfactory bandit regret performance.

Across contextual linear, multimodal and outlier-contaminated bandits, \texttt{Nonparametric TS} empirically recovers Oracle-level performance without knowing the true reward model.
Additionally, it reduces average regret relative to a broad suite of state-of-the-art alternatives, also for non-exponential-family rewards, with regret volatility among the lowest.
It remains computationally practical under a bounded inference budget,
and achieves all of the above with a single algorithm and no per-scenario tuning.

We see three natural directions as follow-up work:
one is empirically evaluating the Pitman-Yor process (power-law tail behavior) and hierarchical modeling (across-arm learning) extensions,
as introduced in Appendices~\ref{asec:pitman_yor} and~\ref{asec:hierarchical}, respectively;
another is extending the per-arm reward model to nonlinear context-reward mean functions (Appendix~\ref{assec:nonparametric_ts_expressivity});
and a third, a formal regret analysis complementing the empirical evidence we presented.

\bibliographystyle{apalike}
\bibliography{literature}

\clearpage
\appendix
\thispagestyle{empty}

\onecolumn
\aistatstitle{
	Bayesian Nonparametric Thompson Sampling \\ 
	for Multi-Armed Bandits: 
	Supplementary Materials
}

\section{Nonparametric TS: details}
\label{asec:nonparametric_ts_details}
\subsection{Model expressivity}
\label{assec:nonparametric_ts_expressivity}

For non-contextual bandits ($x_t=1$ for all $t$), Equation~\eqref{eq:per_arm_dp_observations} defines a DP Gaussian mixture per-arm,
for which the posterior consistency results for density estimation discussed in Section~\ref{sec:background} apply~\citep{j-Ghosal1999,j-Ghosal2007,j-Ghosal2010}.
For contextual bandits, it defines a DP mixture of linear regressions with context-independent mixture weights $\pi_{a,k}$, so that
\begin{equation}
	\mathbb{E}\left[Y|x,G_a\right] = \sum_{k} \pi_{a,k}\, x^\top w_{a,k} = x^\top \Big(\sum_{k} \pi_{a,k}\, w_{a,k}\Big) \;,
\end{equation}
\ie the conditional reward mean is linear in the context,
while the shape of the conditional reward density can be multimodal, skewed or outlier-contaminated.
Posterior consistency results for general conditional densities~\citep{j-Tokdar2006,j-Pati2013} instead rely on modeling covariate dependence, \eg jointly over contexts and rewards, or via covariate-dependent mixture weights as in dependent DPs~\citep{tr-MacEachern2000},
which can also capture nonlinear context-reward mean functions.
In this work, we illustrate the reward density estimation flexibility of \texttt{Nonparametric TS} in the standard contextual linear setting;
extensions to nonlinear mean functions, via nonlinear context features or covariate-dependent mixture weights, are left as future work.

\subsection{Closed-form posterior updates}
\label{assec:nonparametric_ts_posterior}

Due to the conjugacy of the normal inverse-gamma base measure $G_{a,0}$ to the context-conditional Gaussian emission of Equation~\eqref{eq:per_arm_dp_observations}, the posterior of each distinct mixture component's parameters $\varphi_{a,k}^*=(w_{a,k}^*,\sigma_{a,k}^{*2})$, $k=1,\dots,K_{t,a}$, is itself normal inverse-gamma,
\begin{align}
G_{a,n_{t,a,k}}(\varphi_{a,k}^*) = \N{w_{a,k}^*|U_{a,k,n_{t,a,k}},\sigma_{a,k}^{*2}V_{a,k,n_{t,a,k}}} \times \IG{\sigma_{a,k}^{*2}|\alpha_{a,k,n_{t,a,k}},\beta_{a,k,n_{t,a,k}}} \;,
\label{eq:posterior_per_mixture}
\end{align}
with hyperparameters, for $R_{a,k,n_{t,a}} \in \Real^{n_{t,a}\times n_{t,a}}$ a diagonal indicator matrix selecting the $n_{t,a,k}$ observations of arm $a$ assigned to component $k$,
and $x_{(t_a)} \in \Real^{d \times n_{t,a}}$ a concatenation of $n_{t,a}$ context column-vectors,
\begin{align}
\begin{cases}
V_{a,k,n_{t,a,k}}^{-1} &= x_{(t_a)} R_{a,k,n_{t,a}} x_{(t_a)}^\top + V_{a,0}^{-1} \;,\\
U_{a,k,n_{t,a,k}} &= V_{a,k,n_{t,a,k}} \left( x_{(t_a)} R_{a,k,n_{t,a}} y_{(t_a)} + V_{a,0}^{-1} U_{a,0}\right) \;, \\
\alpha_{a,k,n_{t,a,k}} &= \alpha_{a,0} + \tfrac{1}{2} \tr{R_{a,k,n_{t,a}}} \;, \\
\beta_{a,k,n_{t,a,k}} &= \beta_{a,0} + \tfrac{1}{2}\left(y_{(t_a)}^\top R_{a,k,n_{t,a}}y_{(t_a)} \right) 
+ \tfrac{1}{2}\left( U_{a,0}^\top V_{a,0}^{-1} U_{a,0} - U_{a,k,n_{t,a,k}}^\top V_{a,k,n_{t,a,k}}^{-1} U_{a,k,n_{t,a,k}} \right) \;.
\end{cases}
\label{eq:posterior_hyperparameters}
\end{align}

The posterior updates depend on the number of rewards $n_{t,a,k}$ after playing arm $a$ that are assigned to mixture component $k$, \ie
$n_{t,a,k}=\sum_{t^\prime \in (t_a)} \mathds{1}[z_{t^\prime}=k]$.
These updates are computed sequentially as data are observed for the played arm, and are used both in the predictive posterior of Equation~\eqref{eq:nonparametric_Gaussian_mixture_detailed} and in the Gibbs sampler of Algorithm~\ref{alg:nonparametric_gibbs}.

\subsection{BNP Inference: per-arm DP mixture model posterior}
\label{assec:nonparametric_ts_inference}

For the proposed nonparametric Thompson sampling to operate successfully,
we must compute and draw from per-arm DP mixture model posteriors, for which inference of the unknown latent variables $z_{1:t}$, given observations $y_{1:t}$, is needed.
The Bayesian inference literature provides plenty of alternatives to do so, based on Markov chain Monte Carlo~\citep{j-Neal2000}, sequential Monte Carlo~\citep{j-Fearnhead2004} and variational inference~\citep{ip-Blei2004}.
Any of these techniques that infer the assignment variables and result in density convergence is adequate for execution in line 11 of Algorithm~\ref{alg:nonparametric_ts}.

We here describe and implement a marginalized Gibbs sampler ---equivalent to Algorithm 3 of~\citet{j-Neal2000}---
that iterates between sampling assignments $z_{1:t}$ and updates to the emission distribution's hyperparameters $\varPhi_{a,k,n_{t,a,k}}$.
The pseudo-code is presented in Algorithm~\ref{alg:nonparametric_gibbs_full} and is described hereafter.

\paragraph*{Marginalized Gibbs sampler: overview and Algorithm.}
As a new reward $y_{t,a_{t}}$ is observed for the arm $a_t$ played at time $t$,
the BNP mixture model is updated based on the Gibbs sampling procedure of Algorithm~\ref{alg:nonparametric_gibbs_full}.

The Gibbs sampler consists of updating the conditional assignments of each observed reward to its corresponding mixture component.
We use auxiliary latent variables $z_{1:t}$ that keep track of which distinct mixture component each observation $y_{1:t}$ is drawn from.

We iterate through $(t_a)=(\tau\leq t: a_\tau=a)$ to determine the assignments $z_{(t_a)}$ of arm $a$'s rewards to distinct, already instantiated, mixture components $k=(1,\cdots,K_{t,a})$;
or a new mixture component $k_{new}$ ---Line 7 in Algorithm~\ref{alg:nonparametric_gibbs_full}.
The Gibbs sampler autonomously adjusts the posterior's complexity (\ie number of mixture components $K_{t,a}$) according to the observed per-arm rewards.
Given these inferred assignments $z_{(t_a)}$, we recompute sufficient statistics $n_{t,a,k}$ ---Line 8 in Algorithm~\ref{alg:nonparametric_gibbs_full}--- and update parameter posteriors $\varPhi_{a,k,n_{t,a,k}}$ ---Line 9 in Algorithm~\ref{alg:nonparametric_gibbs_full}.

\begin{algorithm}[!h]
	\caption{Gibbs sampler for DP Gaussian mixture posteriors per-arm}
	\label{alg:nonparametric_gibbs_full}
	\begin{algorithmic}[1]
		\STATE {\bfseries Input:} Hyperparameters for arm $a$: $\gamma_a$, $\varPhi_{a,0}$
		\STATE {\bfseries Input:} Observed contexts and rewards for arm $a$: $(x_{(t_a)}, y_{(t_a)})$
		\STATE {\bfseries Input:} Gibbs convergence criteria $\epsilon$, $Gibbs_{max}$
		\STATE Initialize: sample the newest reward's assignment $z_t$ as in line 7, compute $l(1)$ as in line 12, and set $Gibbs_{iter}=1$, $converged=\text{false}$\\
		\WHILE{$Gibbs_{iter}<Gibbs_{max}$ and not $converged$}
		\FOR {$t^\prime \in (t_a)$}
		\STATE Draw mixture assignment for $z_{t^\prime}\in(1,\cdots, K_{t,a}, k_{new})$, \\
		$\hspace*{2ex}$ with hyperparameters $\varPhi_{a,k,n_{t,a,k}}$ as in Equation~\eqref{eq:posterior_hyperparameters}:\\
		$\hspace*{4ex} p(z_{t^\prime}=k|y_{t^\prime},x_{t^\prime}, \varPhi_{a,k,n_{t,a,k}}) \propto 
		\frac{n_{t,a,k}}{n_{t,a}+\gamma_a} \; \T{Y_{t^\prime}|\varPhi_{a,k,n_{t,a,k}}}$, %
		for $k=(1,\cdots,K_{t,a})$\\
		$\hspace*{4ex} p(z_{t^\prime}=k_{new}|y_{t^\prime},x_{t^\prime}, \varPhi_{a,0}) \propto 
		\frac{\gamma_a}{n_{t,a}+\gamma_a} \; \T{Y_{t^\prime}|\varPhi_{a,0}}$. \\
		\STATE Update sufficient statistics\\
		$\hspace*{2ex} n_{t,a,k}=\sum_{t^\prime \in (t_a)} \mathds{1}[z_{t^\prime}=k]\;,$ $\hspace*{2ex}$ for $k=(1,\cdots,K_{t,a})$.
		\STATE Update parameter posteriors $\varPhi_{a,k,n_{t,a,k}}$ as in Equation~\eqref{eq:posterior_hyperparameters}.
		\ENDFOR
		\STATE{Update number of iterations $Gibbs_{iter}$++}
		\STATE Compute joint reward and assignment log-likelihood\\
		$\hspace*{2ex} l(Gibbs_{iter}) = \log p(y_{(t_a)},z_{(t_a)}|\varPhi_{a,n_{t,a}})$ \\  
		\STATE $converged = \left|\frac{l(Gibbs_{iter})-l(Gibbs_{iter}-1)}{l(Gibbs_{iter}-1)}\right|<\epsilon$
		\ENDWHILE
	\end{algorithmic}
\end{algorithm}

The Gibbs sampler is run until a stopping criterion is met ---Line 13 in Algorithm~\ref{alg:nonparametric_gibbs_full}:
either the joint log-likelihood $p(y_{(t_a)},z_{(t_a)}|\varPhi_{a,n_{t,a}})$ of the observed rewards $y_{(t_a)}$ and sampled assignments $z_{(t_a)}$, evaluated at the updated posterior hyperparameters $\varPhi_{a,n_{t,a}}$, is stable within an $\epsilon$ relative log-likelihood margin between consecutive iterations,
or a maximum number of iterations $Gibbs_{max}$ is reached.

\paragraph*{Marginalized Gibbs sampler details: BNP computations.}
Because we are using independent DP mixture models per-arm,
we keep track of per-arm observations $y_{(t_a)}$ and assignments $z_{(t_a)}$, each of size $n_{t,a}$, for a total of $t=\sum_{a=1}^A n_{t,a}$.
These $z_{(t_a)}$ are $K_{t,a}+1$ dimensional categorical variables, where $z_{t^\prime}=k$ for $t^\prime \in (t_a)$ if observation $y_{t^\prime}$ is drawn from arm $a$ and mixture component $k$.

We update these via Gibbs sampling using the following conditional probabilities,
where we denote with $y_{t^\prime}$ a new observation, not in $y_{(t_a)}$,
\begin{align}
	p(z_{t^\prime}&=k|y_{t^\prime},x,y_{(t_a)},z_{(t_a)}, \gamma_a, G_{a,0})
		\propto \frac{n_{t,a,k}}{(n_{t,a}+\gamma_a)} \int_{\varphi_{a,k}^*} p(y_{t^\prime}|x,\varphi_{a,k}^*) \dd{G_{a,n_{t,a,k}}(\varphi_{a,k}^*)} \label{eq:nonparametric_mixture_assignment_seen} \;, \quad  k=1, \cdots, K_{t,a} \;, \\
	p(z_{t^\prime}&=k_{new}|y_{t^\prime},y_{(t_a)},z_{(t_a)}, \gamma_a, G_{a,0} )
	\propto
	\frac{\gamma_a}{(n_{t,a}+\gamma_a)} \int_{\varphi_{a,k_{new}}} p(y_{t^\prime}|x,\varphi_{a,k_{new}}) \dd{G_{a,0}(\varphi_{a,k_{new}})} \;,
	\label{eq:nonparametric_mixture_assignment_new}
\end{align}
and $G_{a,n_{t,a,k}}(\varphi_{a,k})$ is the posterior measure after observing $n_{t,a,k}$ rewards from arm $a$ of per-arm parameters $\varphi_{a,k}$ assigned to distinct mixtures $k=1, \cdots, K_{t,a}$;
while $G_{a,0}(\varphi_{a,k_{new}})$ denotes the per-arm DP prior base measure.
For the proposed prior-likelihood combination,
the posterior $G_{a,n_{t,a}}(\varphi_{a,k})$ can be computed in closed form, as detailed in Equation~\eqref{eq:posterior_hyperparameters}.
Note that when running the Gibbs sampler, we recompute assignment probabilities for existing observations per-arm $a$ as well.
In this case,
for ${t^\prime} \in (t_a)$, one must carefully update counts ($n_{t,a}-1$, $n_{t,a,z_{t^\prime}=k}-1$) and remove observation $y_{t^\prime}$ from all updated posterior hyperparameters $\varPhi_{a,n_{t,a}-1}$ accordingly.

\paragraph*{Marginalized assignment probabilities.}
In order to compute the assignment probabilities for the Gibbs sampler, we marginalize out the parameters $\varphi_a$ on the right-hand side of Equations~\eqref{eq:nonparametric_mixture_assignment_seen}--\eqref{eq:nonparametric_mixture_assignment_new}, \ie 
\begin{align}
	p(Y_{t}|x,\varPhi_{a}) = \int_{\varphi_{a}} p(Y_{t}|x,\varphi_{a}) \dd{G_{a,n}(\varphi_{a})}\;,
\end{align}
where $G_{a,n}(\varphi_a)$ denotes the normal inverse-gamma posterior after $n$ rewards, \ie the prior $G_{a,0}$ for $n=0$ (a potential new component $k_{new}$), or the posterior $G_{a,n_{t,a,k}}(\varphi_{a,k}^*)$ for distinct mixture components $k=1, \cdots, K_{t,a}$.

Due to the conjugacy of the emission distribution and $G_{a,n}(\varphi_a)$ ---a normal inverse-gamma distribution for both the prior and the posterior---
the per-arm reward predictive emission distribution follows a context conditional Student-t distribution
\begin{align}
	p(Y_{t}|x,\varPhi_{a}) 
	&= \T{Y_{t}|\nu_{a}, m_{a}, r_{a}} \;,
	\qquad 
	\varPhi_{a} =
	\begin{cases}
		\nu_{a}=2\alpha_{a} \;, \\
		m_{a} = x^\top U_{a} \;, \\
		r_{a}^2 = \frac{\beta_{a}}{\alpha_{a}} (1+x^\top V_{a} x) \;.
	\end{cases}
	\label{eq:marginalized_predictive_emission_univariate}
\end{align}

The hyperparameters $\varPhi_{a}=\{\nu_{a}, m_{a}, r_{a}\}$ are computed based on the prior $\varPhi_{a,0}$, or the posterior $\varPhi_{a,k,n_{t,a,k}}$ as in Equation~\eqref{eq:posterior_hyperparameters}, 
depending on whether the predictive density refers to 
a `\textit{new}' component $k=k_{new}$ with $n_{t,a,k_{new}}=0$ in Equation~\eqref{eq:nonparametric_mixture_assignment_new},
or a
`\textit{distinct}' already instantiated components $k=1,\cdots,K_{t,a}$ in Equation~\eqref{eq:nonparametric_mixture_assignment_seen}, for which $n_{t,a,k}\geq1$ observations have been already assigned to, respectively.

The likelihood of a set of rewards assigned to per-arm distinct component $k$, $y_{(t_{a,k})}=y_{1:t}\cdot \mathds{1}[a_t=a,z_t=k]$, given their associated contexts $x_{(t_{a,k})}=x_{1:t} \cdot \mathds{1}[a_t=a,z_t=k]$, with the component's parameters marginalized with respect to the normal inverse-gamma $G_{a,n}$ (which will be of use in the Gibbs sampler's convergence diagnostic) obeys the following matrix t-distribution
\begin{align}
	p(Y_{(t_{a,k})}|x_{(t_{a,k})},\varPhi_{a,k,n_{t,a,k}})
	& = \MT{Y_{(t_{a,k})}|\nu_{a,k,n_{t,a,k}}, M_{a,k,n_{t,a,k}}, \Psi_{a,k,n_{t,a,k}}, \Omega_{a,k,n_{t,a,k}}} \; , \nonumber  \\
	\text{with }& \begin{cases}
		\nu_{a,k,n_{t,a,k}}=2 \alpha_{a,k,n_{t,a,k}} \;,\\
		M_{a,k,n_{t,a,k}}= x_{(t_a),k}^\top U_{a,k,n_{t,a,k}} \;, \\
		\Psi_{a,k,n_{t,a,k}} = I_{n_{t,a,k}} + x_{(t_a),k}^\top V_{a,k,n_{t,a,k}} x_{(t_a),k} \;, \\
		\Omega_{a,k,n_{t,a,k}} = 2 \beta_{a,k,n_{t,a,k}} \;.
	\end{cases}
	\label{eq:marginalized_predictive_emission_multivariate}
\end{align}

\paragraph*{Convergence diagnostic: joint data and assignment log-likelihood.}
A key step of the Gibbs sampler is to determine its convergence criteria,
for which we monitor the joint likelihood of per-arm observations and nonparametric assignments,
with each component's reward likelihood evaluated at its current posterior hyperparameters $\varPhi_{a,k,n_{t,a,k}}$:
a deterministic function of the current assignments $z_{(t_a)}$, which stabilizes as these do.
This joint likelihood is computed via the following factorization
---note that the following is written in a general form, applying to PY processes; for the DP, $d_a=0$:
\begin{equation}
	p(y_{(t_a)},z_{(t_a)}|\varPhi_{a,n_{t,a}}) = p(y_{(t_a)}|z_{(t_a)}, \varPhi_{a,n_{t,a}}) p(z_{(t_a)}|d_a, \gamma_a) \; ,
	\label{eq:nonparametric_mixture_data_assignment_likelihood}
\end{equation}
where the conditional data-likelihood term
\begin{align}
	p(y_{(t_a)}|z_{(t_a)}, \varPhi_{a,n_{t,a}})
	& = \prod_{k=1}^{K_{t,a}} p(y_{(t_{a,k})}|z_{(t_{a,k})},\varPhi_{a,k,n_{t,a,k}})  \;,
\end{align}
is computed as shown in Equation~\eqref{eq:marginalized_predictive_emission_multivariate}.
The prior probability of the auxiliary assignment variables $z_{(t_a)}$ factorizes as follows:
\begin{equation}
	p(z_{(t_a)}|d_a, \gamma_a) = \prod_{i\in (t_a)} p(z_{i}|z_{1:i-1},d_a, \gamma_a) \; ,
	\label{eq:nonparametric_mixture_assignment_likelihood}
\end{equation}
which allows for computation of the joint distribution ---details can be found in~\citet{j-Buntine2010}--- of the $n_{t,a}$ assignments to distinct mixture components $k$ per-arm:
\begin{equation}
	p(z_{(t_a)}|d_a, \gamma_a) = \frac{ [\gamma_a | d_a]^{(K_{t,a})}}{[\gamma_a]^{(n_{t,a})}}\prod_{k=1}^{K_{t,a}}  [1-d_a]^{(n_{t,a,k}-1)} \; .
	\label{eq:py_mixture_assignment_likelihood}
\end{equation}
$[\gamma_a]^N$ above denotes the rising factorial (also known as the Pochhammer function or polynomial),
$[\gamma_a]^N =\gamma_a \cdot (\gamma_a + 1) \cdots (\gamma_a+N-1)$;
and
$[\gamma_a | d_a]^{(K)}$ denotes the Pochhammer polynomial with increment $d_a$:
$[\gamma_a | d_a]^{(K)} = \gamma_a \cdot (\gamma_a + d_a) \cdots (\gamma_a+(K-1)d_a)$.

\paragraph{Warm-started Gibbs sampler.}
Due to the sequential acquisition of rewards in the bandit setting ---a single arm is played at each bandit interaction--- only the posterior for the played arm $a_t=a$ needs to be updated at time $t$.
As such, even though the Gibbs sampler might be restarted with prior hyperparameters $G_{a,0}$ at every interaction,
we propose to instead initialize the latent assignments $z_{(t_a)}$ and posterior hyperparameters $\varPhi_{a,n_{t,a}-1}$ from the previous step,
while leaving the base measure $G_{a,0}$ fixed.
Namely, we set the DP hyperparameters and inferred assignments for all but this latest observed reward as input to Algorithm~\ref{alg:nonparametric_gibbs},
\ie with $\varPhi_{a,n_{t,a}-1}$ as inferred based on the already observed $n_{t,a}-1$ rewards.
Consequently, the Gibbs sampler is \textit{warm-started} at each bandit interaction with good hyperparameters and latent assignments (that describe all but this newly observed reward $y_{t,a_{t}}$), so that convergence can be achieved in few iterations per observed reward.

\paragraph{Computational complexity.}
\label{sssec:nonparametric_thompson_sampling_computational_complexity}

Since Equation~\eqref{eq:posterior_hyperparameters} can be sequentially computed for each per-arm observation, the computational cost of the Gibbs sampler in Algorithm~\ref{alg:nonparametric_gibbs} grows with the number of rewards observed for the played arm $n_{t,a}$, with $t=\sum_{a=1}^{A}n_{t,a}$.
Therefore, the overall computational cost is upper-bounded by $\mathcal{O}(Gibbs_{iters}\, n_{t,a}\,(K_{t,a}+1))$ per-interaction with the world, \ie per newly observed reward $y_{t,a_{t}}$.
This bound counts the predictive-density evaluations of the assignment step:
each Gibbs iteration revisits the arm's $n_{t,a}$ rewards, and scores $K_{t,a}+1$ components (all instantiated ones and a potential new one) for each,
\ie at most $\mathcal{O}(Gibbs_{iters}\, T^2 \max_{t,a} K_{t,a})$ over $T$ interactions.
Since the expected number of components under the DP prior grows only logarithmically with the number of observations, $\mathbb{E}[K_{t,a}]\approx\gamma_a\log(1+n_{t,a}/\gamma_a)$~\citep{j-Antoniak1974},
the expected per-interaction cost is a priori $\mathcal{O}(Gibbs_{iters}\, n_{t,a}\log n_{t,a})$.
In addition, our implementation of the convergence diagnostic of Equation~\eqref{eq:marginalized_predictive_emission_multivariate} involves determinants of $n_{t,a,k}\times n_{t,a,k}$ matrices per component and iteration,
which could be reduced to $d\times d$ determinants via the matrix determinant lemma.

In practice, and because of the \textit{warm-start}, one can limit the number of Gibbs sampler iterations per-bandit interaction to upper-bound the algorithm's complexity to $\mathcal{O}(Gibbs_{max}\, n_{t,a}\,(K_{t,a}+1))$ per MAB interaction with the environment\footnote{
	Note that, as reported by \citet{j-Fearnhead2004}, the computational cost of running a Gibbs sampler for $I$ iterations is similar to running a sequential Monte Carlo method with $I$ particles.},
yet achieve satisfactory performance ---empirical evidence of this claim is provided in Appendix~\ref{asec:gibbs}.
We emphasize that the Gibbs sampler shall run until log-likelihood convergence, 
and only suggest limiting the number of Gibbs iterations to $Gibbs_{max}$ as a practical recommendation with upper-bounded computational complexity and good empirical regret performance.

\clearpage
\section{Nonparametric TS: full experiments and details}
\label{asec:nonparametric_ts_experiments}

Code to reproduce all experiments, with instructions and environment specifications, is available at \url{https://anonymous.4open.science/r/nonparametric_ts-688D}.
\subsection{Baseline descriptions and hyperparameters}
\label{asec:baselines}

We compare \texttt{Nonparametric TS} as in Algorithm~\ref{alg:nonparametric_ts} with independent DP priors (see details below) against the following Thompson sampling baselines
---note that a full description of these baselines and their implementation can be found in~\citet{deep_bayesian_bandits} (Apache 2.0 license)---
which are state-of-the-art techniques that provide different computations of (or approximations to) reward posteriors that can be combined with Thompson sampling to address complex contextual bandits.

\begin{enumerate}
	\item \texttt{Nonparametric TS}: we use independent Dirichlet process priors per arm (\ie Pitman-Yor discount $d_a=0$)
	with concentration $\gamma_a=0.1$, and a normal-inverse-gamma base measure with
	$\alpha_{a,0}=\beta_{a,0}=1$, $U_{a,0}=\mathbf{1}$ (a $d$-dimensional vector of ones) and $V_{a,0}=I$,
	i.e., $\sigma_{a,k}^2\sim\mathcal{IG}(1,1)$ and $w_{a,k}|\sigma_{a,k}^2\sim\N{\mathbf{1},\sigma_{a,k}^2 I}$.
	Each round, \texttt{Nonparametric TS} draws a single posterior sample per arm, computing the expected reward with the expected mixture weights of Eq.~\eqref{eq:dp_expected_reward}.
	The warm-started Gibbs sampler stops when the relative change in the joint log-likelihood falls below $\epsilon=10^{-4}$, or after $Gibbs_{max}=10$ iterations.
	
	\item \texttt{LinearGaussian TS}: A conjugate, non-neural Thompson sampling baseline that assumes a contextual linear Gaussian reward function,
	$
	Y_{t,a}\sim \N{Y|\theta_a^\top x_t, \sigma_a^2}. %
	$
	In its simplest setting, when the reward variance is known, the resulting Thompson sampling implementation follows that by~\citet{ip-Agrawal2013a}. In our experiments, in a similar fashion as done by~\citet{ip-Riquelme2018}, we model the joint distribution of $\theta_a$ and $\sigma_a^2$, $\forall a \in \A$, which allows the method to adaptively adjust to the observed reward noise. We leverage the normal inverse-gamma conjugate prior, \ie
	\begin{equation}
		\left(\theta_a, \sigma_a^2\right) \sim \N{\theta_a |U_{a,0}, \sigma_a^2V_{a,0}} \IG{\sigma_a^2 |\alpha_{a,0}, \beta_{a,0}} \; ,
	\end{equation}
	of the Gaussian contextual model to derive the exact Bayesian posterior.
	
	In the linear Gaussian Oracle comparisons (Section~\ref{ssec:rq1}),
	we use our own implementation, without intercept, 
	with prior $w_a|\sigma_a^2\sim\N{\mathbf{1},\sigma_a^2 I}$, i.e., the same prior as each mixture component of \texttt{Nonparametric TS}.
	In the rest of experiments, we directly use the \texttt{LinFullPost} implementation of~\citet{ip-Riquelme2018},
	which includes an intercept term, with prior $w_a|\sigma_a^2\sim\N{\mathbf{0},\sigma_a^2\lambda^{-1}I}$ and $\lambda=1$.
	
	\item \texttt{MultitaskGP}: This is an alternative and popular Bayesian nonparametric technique that models context to reward mappings via Gaussian processes~\citep{ip-Srinivas2010,ip-Gruenewaelder2010,ic-Krause2011}. This implementation regresses the expected reward of different context-actions pairs by fitting a multi-task Gaussian process given observed bandit data.
	
	\item \texttt{NeuralLinear}: This algorithm, introduced by~\citet{ip-Riquelme2018}, operates by learning a neural network that maps contexts to rewards for each action, and simultaneously, updates a Bayesian linear regression in the network's last layer that maps a learned representation $h$ linearly to the rewards $y$. The corresponding Thompson sampling draws the learned linear regression parameters $\theta_a$ for each arm, but keeps the representation $h$ output by the learned network. In our experiments, the representation network and the Bayesian linear posterior are retrained and updated at every MAB interaction.
	
	\item \texttt{NeuralBootstrapped}: This algorithm is based on the work by~\citet{ic-Osband2016}, which trains simultaneously and in parallel $Q$ neural networks based on different bootstrapped bandit histories $\HH_{1:t}^{(1)},\cdots, \HH_{1:t}^{(Q)}$. These are generated by adding each newly observed evidence $(x_{t}, a_{t}, y_{t})$ to each history $\HH_{1:t-1}^{(q)}$, $q=(1, \cdots,Q)$, independently and with probability $p \in (0, 1]$. In order to choose an action for a given context, one of the $Q$ networks is selected with uniform probability ($1/Q$), and the best action according to the selected network is played.
	
	\item \texttt{NeuralRMS}: This is a bandit benchmark that trains a neural network to map contexts to rewards. At each time $t$, it acts $\epsilon$-greedily according to the current model, which due to the Stochastic Gradient Descent (SGD) algorithm used for training (the RMSProp optimizer in this implementation), captures randomness in its output.
	
	\item \texttt{NeuralDropoutRMS}: Dropout is a neural network training technique where the output of each neuron is independently zeroed out with a given probability in each forward pass. Once the network is trained, dropout can also be used to obtain a distribution of predictions for a specific input. By choosing the best action with respect to the random dropout prediction, an implicit form of Thompson sampling is executed.
	
	\item \texttt{NeuralParamNoise}: An approach to approximate a distribution over neural networks is to randomly perturb the point estimates attained by SGD on the available data~\citep{ip-Plappert2018}. In this case, the model uses a heuristic to control the amount of \iid noise it adds to the neural network parameters, which is used for making a decision, but not for training: SGD re-starts from the last, noiseless parameter value.
	
	\item \texttt{BNNVariationalGaussian}: This algorithm is based on ideas presented by~\citet{ip-Blundell2015} that combine stochastic variational inference and Bayes by backpropagation. It implements a Bayesian neural network by modeling each individual weight posterior as a univariate Gaussian distribution. Thompson sampling then draws a network at each time step by sampling each weight independently. The variational posterior's parameters are fit by maximizing the evidence lower bound (ELBO) on the observed data.
	
	\item \texttt{BNNAlphaDiv}: This technique leverages expectation-propagation and Black-box alpha-divergence minimization as in~\citet{ip-Hernandez-Lobato2016} to approximate the unknown reward distribution. The algorithm iteratively approximates the posterior of interest by updating a single approximation factor at a time, which usually corresponds to the likelihood of one data point. The implementation adopted here optimizes the global objective directly via stochastic gradient descent.
	
	\item \texttt{Oracle TS}: We implement Oracle Thompson Sampling policies for each evaluated bandit scenario, where the Oracles know the true per-arm reward distributions of the bandit they are targeted to, but need to learn the unknown parameters.
For the mixture and outlier-contaminated scenarios, the Oracle TS uses the finite Gaussian mixture reward model of~\citet{ip-Urteaga2018} with the true number of components $K_a$,
but inferred with the same warm-started Gibbs sampler as \texttt{Nonparametric TS} (rather than via variational inference), with a Dirichlet prior ($\gamma_a=0.1$) on the mixture weights,
so that both policies differ only in whether $K_a$ is known;
the misspecified TS of Section~\ref{ssec:rq1} is the same policy with $K_a=1$.
We detail in Table~\ref{tab:evaluation_scenarios} which state-of-the-art Thompson sampling is appropriate (with references to the original work) for each bandit scenario.
	
	\item \texttt{Optimal}: When possible (\ie for simulated bandits for which there is ground truth), we implement an optimal multi-armed bandit policy that selects the best arm for the given context, based on the known true expected reward of each arm.
\end{enumerate}

We follow the implementations of the \href{https://github.com/tensorflow/models/tree/master/research/deep_contextual_bandits}{Deep Contextual Bandits Showdown}~\citep{ip-Riquelme2018},
and present the set of hyperparameters used for our experiments in Tables below.
Specifically, shared neural network hyperparameters (Table~\ref{tab:neural_network_hyperparameters}, networks are retrained at every bandit interaction), per-baseline details (Table~\ref{tab:neural_network_baselines}), and Gaussian process hyperparameters (Table~\ref{tab:gp_hyperparameters}) follow the suggestions by~\citet{ip-Riquelme2018}, who report that these methods are sensitive to hyperparameter choice in a dataset-dependent way.
As discussed in Section~\ref{sec:discussion}, we did not retune them per scenario. 

\begin{table}[!h]
\caption{Shared neural network hyperparameters.}
\label{tab:neural_network_hyperparameters}
\vspace{-2ex}
\centering
\small
\begin{tabular}{lc}
\toprule
Hyperparameter & Value \\
\midrule
training freq & 1 \\
training epochs & 50 \\
activation & \texttt{tf.nn.relu} \\
layer size & 50 \\
batch size & 512 \\
init scale & 0.3 \\
optimizer & RMS \\
initial pulls & 2 \\
max grad norm & 5.0 \\
initial lr & 0.1 \\
reset lr & True \\
activate decay & True \\
lr decay rate & 0.5 \\
\bottomrule
\end{tabular}
\end{table}

\begin{table}[!h]
\caption{Per-baseline algorithm hyperparameters.}
\label{tab:neural_network_baselines}
\vspace{-2ex}
\centering
\small
\begin{tabular}{p{0.28\linewidth}p{0.62\linewidth}}
\toprule
Algorithm & Details \\
\midrule
\texttt{NeuralLinear} & Network and last-layer posterior updated every iteration; prior $a_0=6,b_0=6,\lambda_0=0.25$. \\
\texttt{NeuralRMS} & RMS optimizer, default parameters. \\
\texttt{NeuralBootstrapped} & $Q=3$ networks/datasets, $p=0.95$. \\
\texttt{NeuralParamNoise} & Noise $\N{0,\sigma=0.05}$; $\epsilon=0.1$-greedy, 300 samples. \\
\texttt{NeuralDropoutRMS} & Dropout probability $0.2$ (keep probability $0.8$), RMS optimizer. \\
\texttt{BNNVariationalGaussian} & Independent Gaussian weight noise, $\sigma=0.1$; 100 initial training steps. \\
\texttt{BNNAlphaDiv} & $\alpha=1$, noise $\sigma=0.1$, $k=20$, prior variance $\sigma^2=0.1$; 100 initial training steps, 20 cleared times. \\
\bottomrule
\end{tabular}
\end{table}

\begin{table}[!h]
\caption{\texttt{MultitaskGP} hyperparameters.}
\label{tab:gp_hyperparameters}
\vspace{-2ex}
\centering
\small
\begin{tabular}{lc}
\toprule
Hyperparameter & Value \\
\midrule
training freq & 1 \\
training epochs & 50 \\
learn embeddings & True \\
task latent dim & 5 \\
max num points & 1000 \\
batch size & 512 \\
optimizer & RMS \\
lr / initial lr & 0.01 / 0.1 \\
lr decay rate & none \\
\bottomrule
\end{tabular}
\end{table}

\clearpage

\subsection{Contextual linear Gaussian bandit: full results}
\label{asec:linear_baselines}

Linear and sparse linear bandits of~\citet{ip-Riquelme2018}: $|\A|=8$ (dense) and $|\A|=7$ (sparse) arms, contexts $x_t\sim\N{0,I_{10}}$, $\theta_a\in\Real^{10}$ (3 non-zero components per arm in the sparse setting), and Gaussian noise with standard deviation $0.01\,a$ for arm $a$.

\begin{figure}[!h]
\centering
\includegraphics[width=0.48\linewidth]{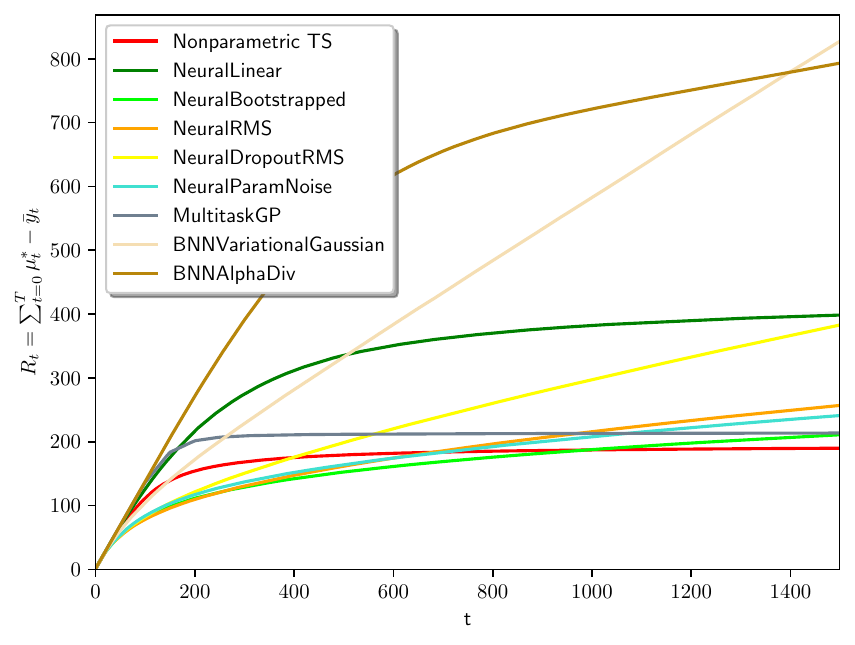}
\includegraphics[width=0.48\linewidth]{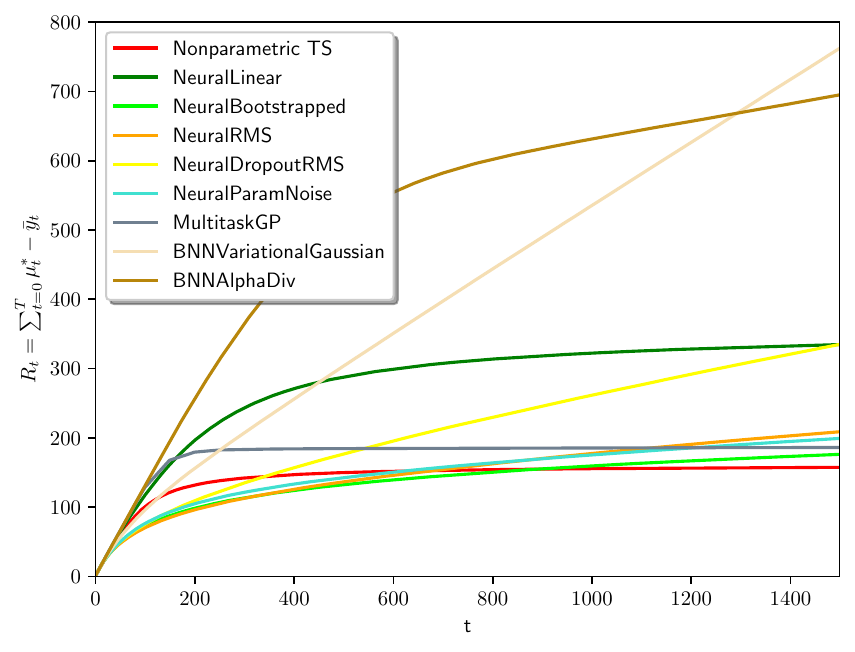}
\caption{Mean $R_t$, dense (left) and sparse (right) linear Gaussian bandit.}
\end{figure}

\begin{table}[!h]
\caption{$R_T$ (mean $\pm$ std) at $T=1500$. Parentheses: relative increase over \texttt{Nonparametric TS}.}
\label{tab:linear_showdown_baselines_regret_full}
\centering
\small
\begin{tabular}{lcc}
\toprule
Algorithm & Linear Gaussian & Sparse Linear Gaussian \\
\midrule
\textbf{Nonparametric TS} & \textbf{189.9 $\pm$ 13.0} (0\%) & \textbf{157.4 $\pm$ 12.4} (0\%) \\
NeuralLinear & 398.5 $\pm$ 18.2 (109.9\%) & 334.6 $\pm$ 16.4 (112.6\%) \\
NeuralBootstrapped & 210.9 $\pm$ 47.4 (11.1\%) & 176.1 $\pm$ 46.6 (11.9\%) \\
NeuralRMS & 256.9 $\pm$ 69.2 (35.3\%) & 208.7 $\pm$ 65.0 (32.6\%) \\
NeuralDropoutRMS & 382.9 $\pm$ 87.1 (101.6\%) & 334.9 $\pm$ 92.3 (112.7\%) \\
NeuralParamNoise & 241.0 $\pm$ 59.3 (26.9\%) & 199.1 $\pm$ 62.1 (26.5\%) \\
MultitaskGP & 213.6 $\pm$ 22.0 (12.5\%) & 186.0 $\pm$ 21.0 (18.2\%) \\
BNNVariationalGaussian & 827.4 $\pm$ 193.1 (335.7\%) & 762.3 $\pm$ 203.5 (384.3\%) \\
BNNAlphaDiv & 793.1 $\pm$ 43.5 (317.7\%) & 695.1 $\pm$ 39.6 (341.6\%) \\
\bottomrule
\end{tabular}
\end{table}

\subsection{Oracle-equivalence sweep}
\label{asec:linear_oracle}

\begin{figure*}[!h]
\centering
\includegraphics[width=0.32\linewidth]{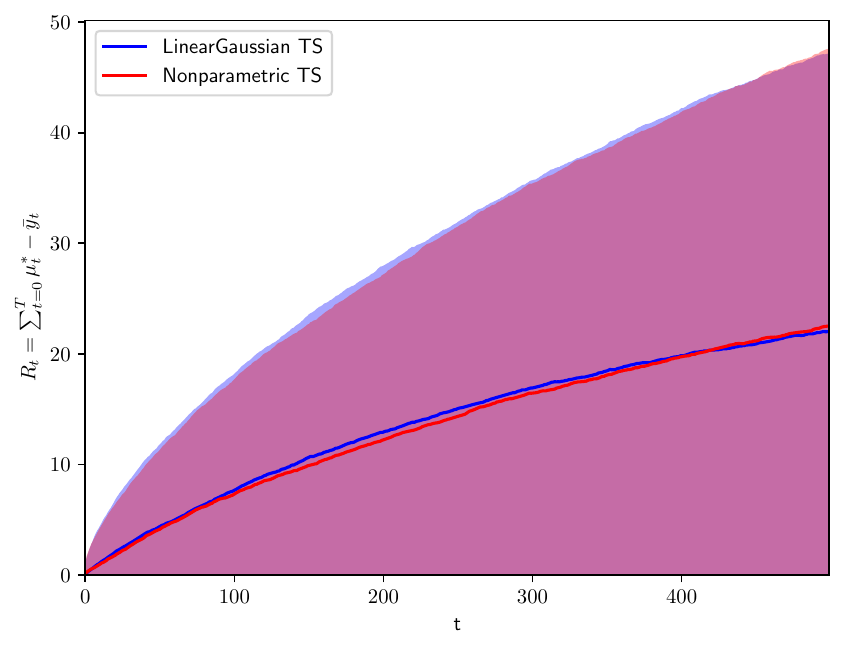}
\includegraphics[width=0.32\linewidth]{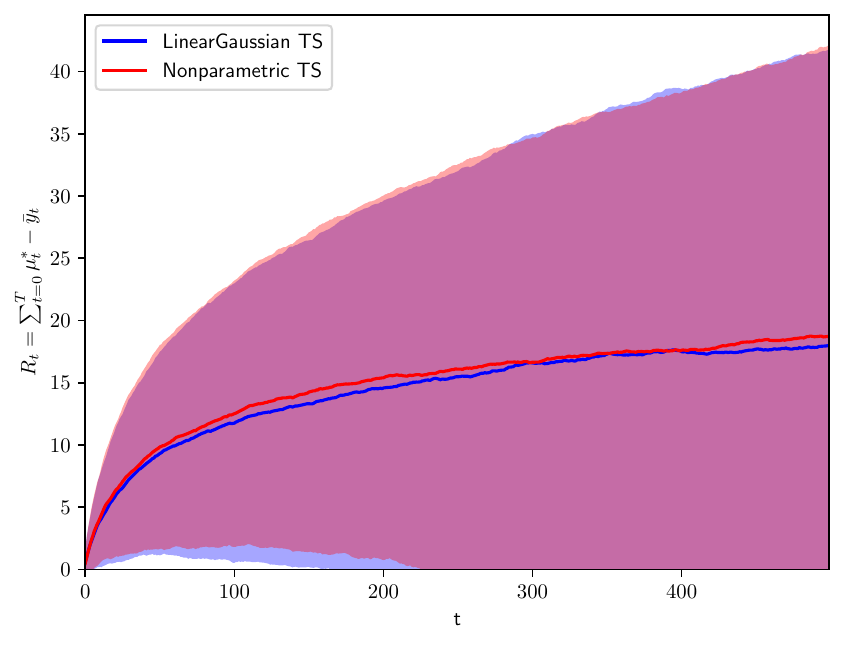}
\includegraphics[width=0.32\linewidth]{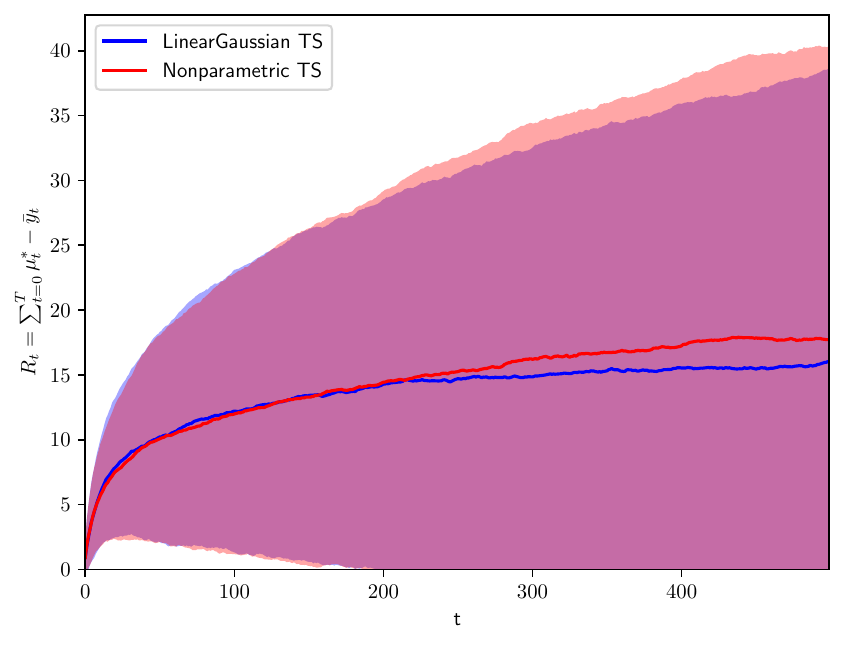}
\includegraphics[width=0.32\linewidth]{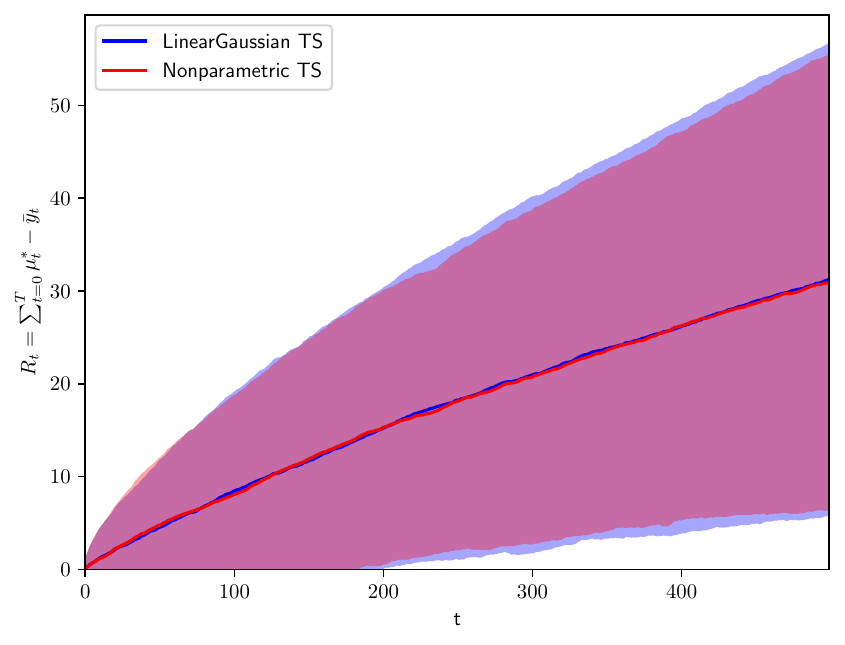}
\includegraphics[width=0.32\linewidth]{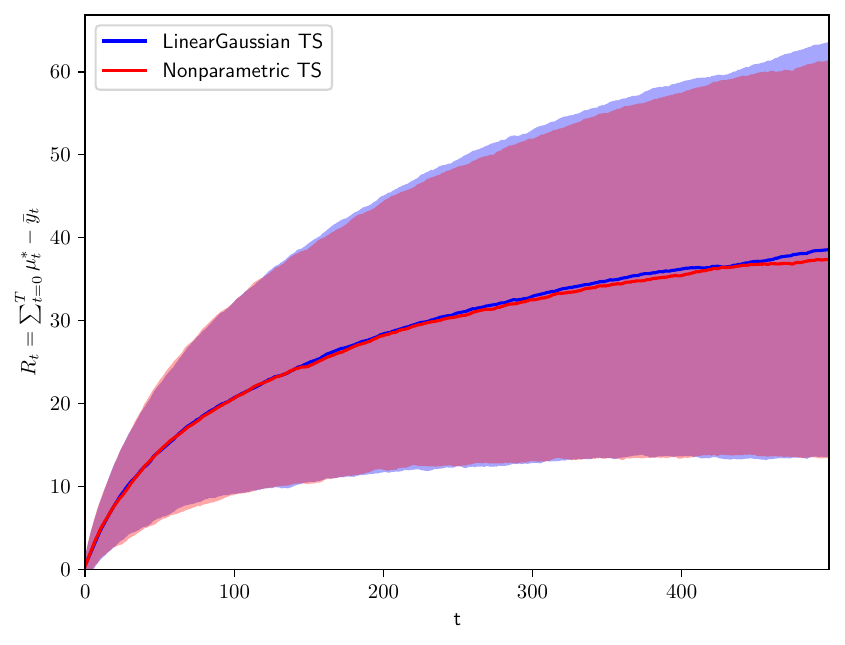}
\includegraphics[width=0.32\linewidth]{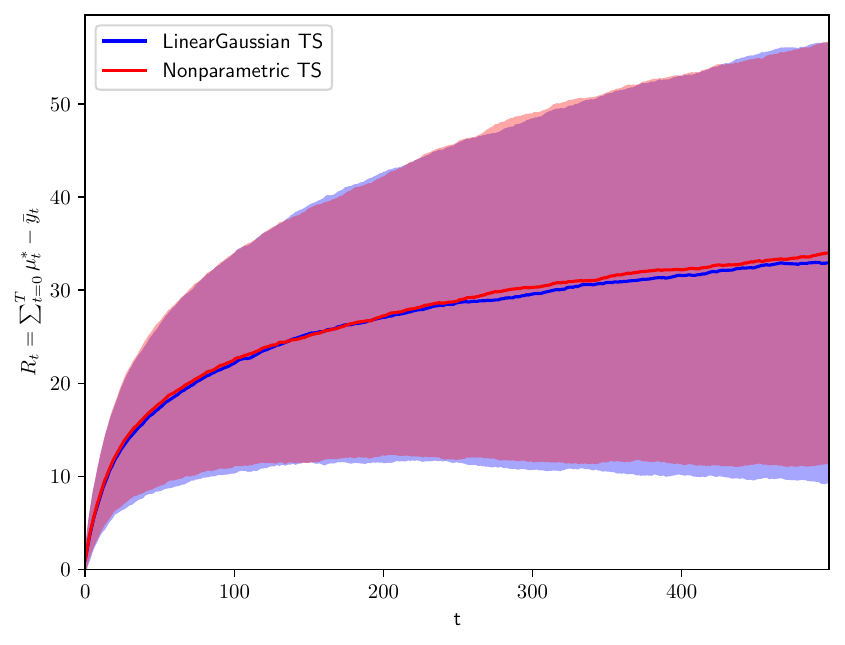}
\includegraphics[width=0.32\linewidth]{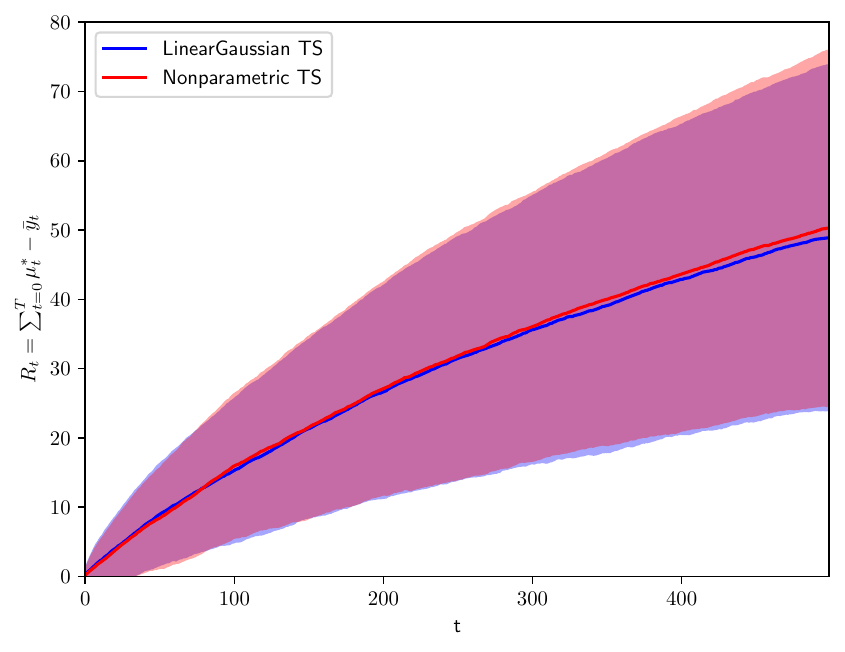}
\includegraphics[width=0.32\linewidth]{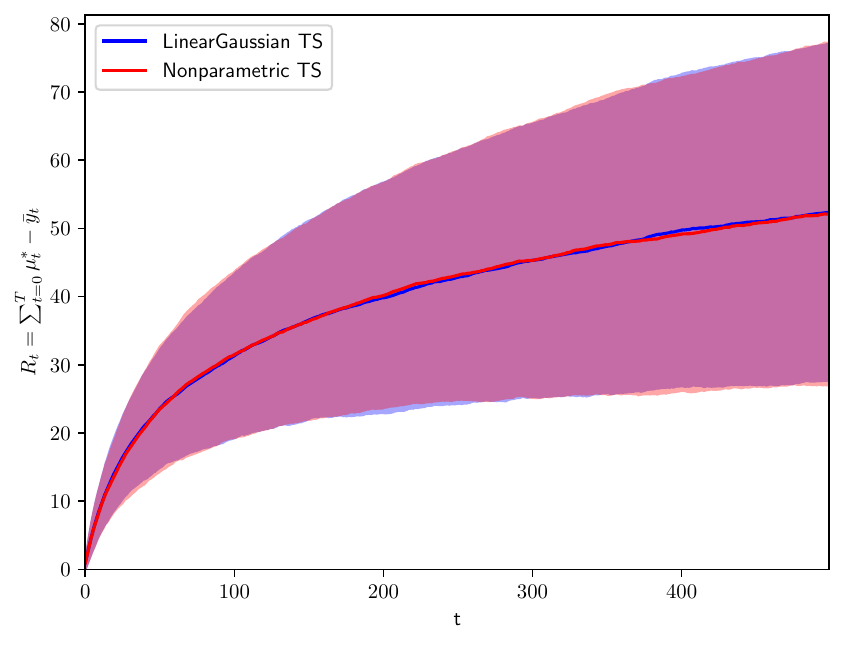}
\includegraphics[width=0.32\linewidth]{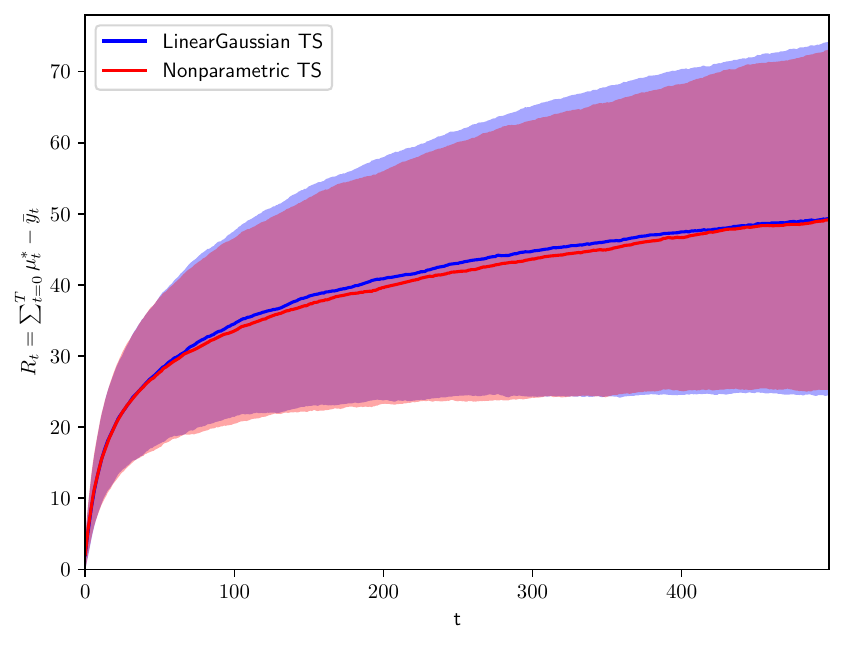}
\includegraphics[width=0.32\linewidth]{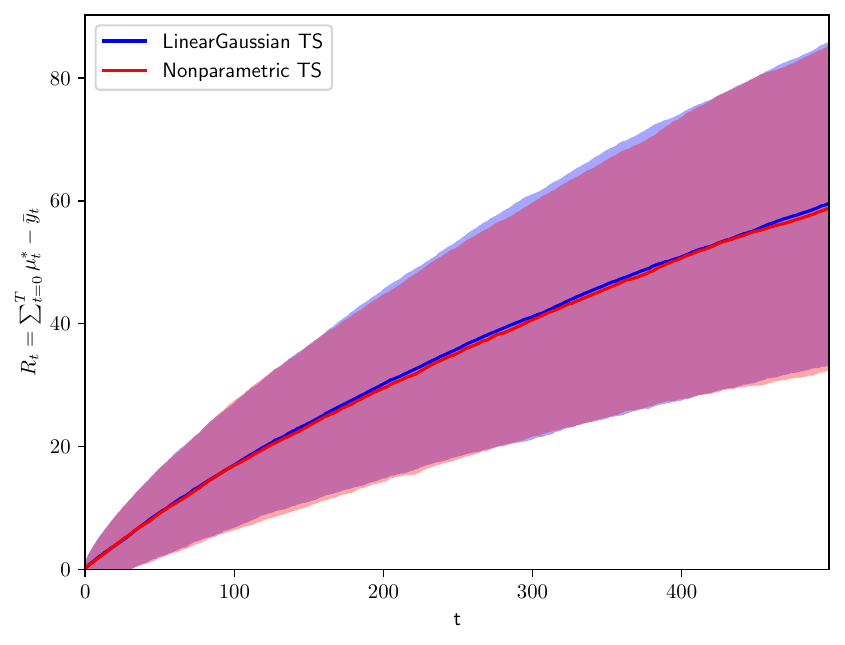}
\includegraphics[width=0.32\linewidth]{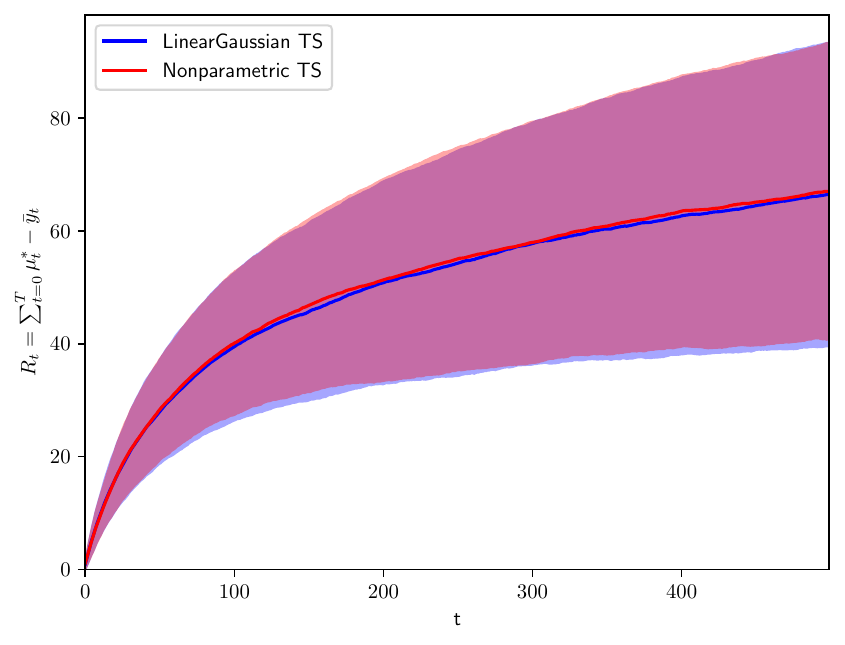}
\includegraphics[width=0.32\linewidth]{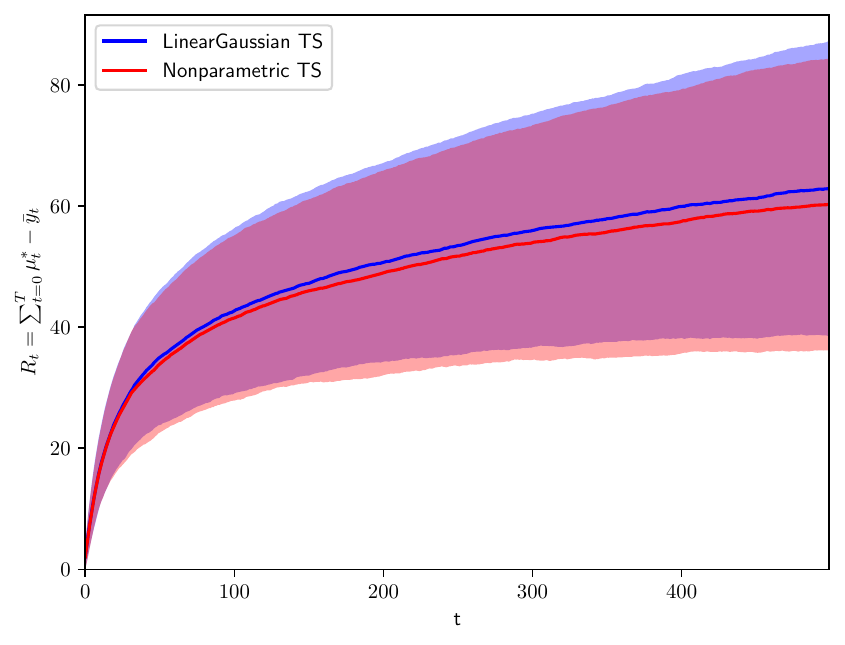}
\caption{Mean $R_t$ (std.\ shaded) for $N=1000$ realizations of $|\A|\in\{2,3,4,5\}$-armed contextual linear Gaussian bandits with varying arm separation, $\sigma_a^2=1\;\forall a$: \texttt{Nonparametric TS} vs.\ \texttt{LinearGaussian TS} (Oracle).}
\label{afig:linear_gaussian_oracle}
\end{figure*}

Figure~\ref{afig:linear_gaussian_oracle} compares \texttt{Nonparametric TS} to the correctly-specified \texttt{LinearGaussian TS}~\citep{ip-Agrawal2013a} across $|\A|\in\{2,3,4,5\}$ armed contextual linear Gaussian bandits with varying arm separation ($N=1000$, two-dimensional context $x_{t}\sim\U{0,1}^2$). Across all parameterizations, \texttt{Nonparametric TS} attains cumulative regret comparable, in both mean and volatility, to the Oracle: the nonparametric predictive posterior converges quickly enough to the true (simple) distribution to incur minimal additional regret.

\clearpage
\subsection{Gaussian mixture bandits: full results}
\label{asec:mixture_baselines}

\texttt{Scenario A}: two-component mixtures per arm, significant overlap between arms, one arm unbalanced ($0.3$/$0.7$ mixing weights).
\texttt{Scenario B}: a \emph{different} reward distribution per arm --- unimodal for arm 1, bimodal for arm 2, an unbalanced three-component mixture for arm 3.
Two-dimensional context $x_{t}\sim\U{0,1}^2$.

\begin{table}[!h]
\caption{$R_T$ (mean $\pm$ std) at $T=1000$. Parentheses: relative increase over \texttt{Nonparametric TS}.}
\label{tab:mixture_scenarios_bandit_showdown_baselines_regret_full}
\centering
\small
\begin{tabular}{lcc}
\toprule
Algorithm & Scenario \texttt{A} & Scenario \texttt{B} \\
\midrule
\textbf{Nonparametric TS} & \textbf{27.5 $\pm$ 70.7} (0\%) & \textbf{44.1 $\pm$ 70.4} (0\%) \\
LinearGaussian TS & 51.2 $\pm$ 99.3 (86.5\%) & 65.4 $\pm$ 89.9 (48.3\%) \\
NeuralLinear & 57.0 $\pm$ 91.6 (107.6\%) & 78.1 $\pm$ 95.0 (77.0\%) \\
NeuralBootstrapped & 137.9 $\pm$ 220.2 (402.0\%) & 166.9 $\pm$ 267.8 (278.4\%) \\
NeuralRMS & 147.0 $\pm$ 227.5 (435.2\%) & 183.3 $\pm$ 282.4 (315.5\%) \\
NeuralDropoutRMS & 143.3 $\pm$ 224.6 (421.7\%) & 182.6 $\pm$ 282.1 (313.8\%) \\
NeuralParamNoise & 70.5 $\pm$ 71.7 (156.6\%) & 103.7 $\pm$ 72.6 (135.1\%) \\
MultitaskGP & 116.2 $\pm$ 99.0 (323.0\%) & 185.7 $\pm$ 152.1 (320.9\%) \\
BNNVariationalGaussian & 126.9 $\pm$ 229.3 (361.9\%) & 173.8 $\pm$ 293.2 (294.0\%) \\
BNNAlphaDiv & 197.4 $\pm$ 66.6 (618.5\%) & 325.7 $\pm$ 68.6 (638.2\%) \\
\bottomrule
\end{tabular}
\end{table}

\begin{figure}[!h]
\centering
\includegraphics[width=0.48\linewidth]{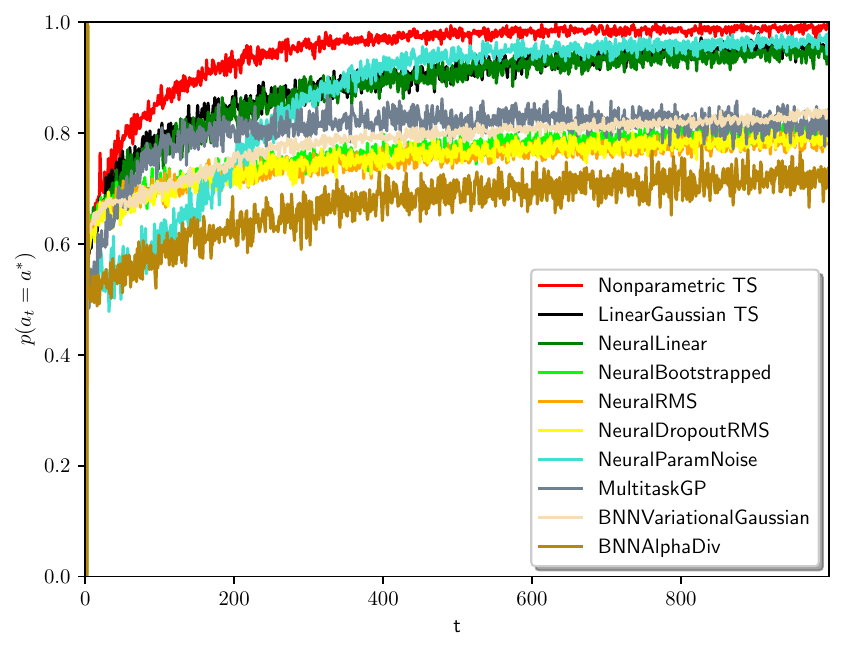}
\includegraphics[width=0.48\linewidth]{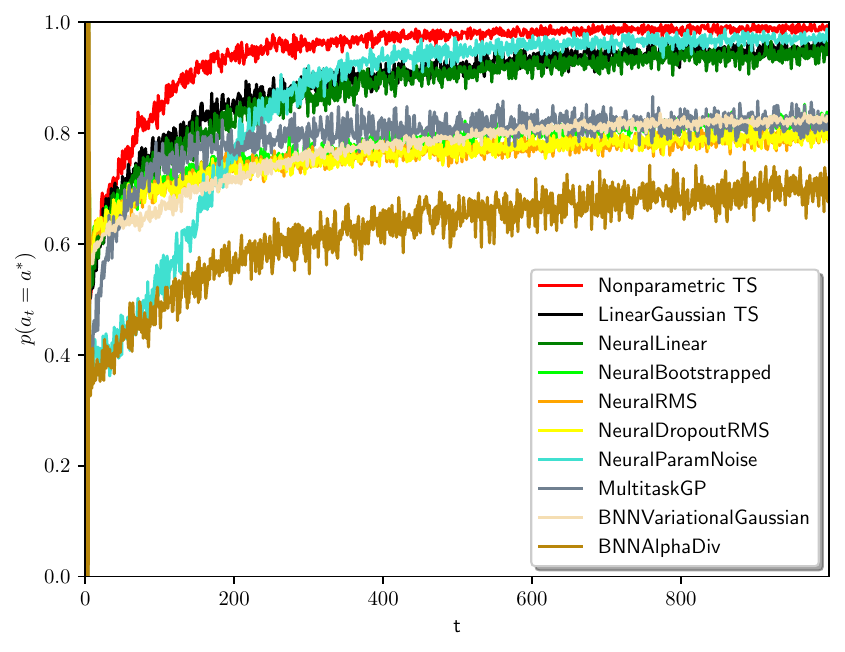}
\caption{Averaged empirical frequency of playing the optimal arm, Scenario \texttt{A} (left) and \texttt{B} (right).}
\end{figure}

\subsection{Outlier-contaminated bandit: full results}
\label{asec:heavy_tail_baselines}

Per-arm rewards are a two-component mixture with shared mean and inflated variance ($\sigma^2=100$ with probability $0.25$), producing outlier-contaminated observations.

\begin{table}[!h]
\caption{$R_T$ (mean $\pm$ std) at $T=1000$. Parentheses: relative increase over \texttt{Nonparametric TS}.}
\label{tab:heavy_tailed_baselines_regret_full}
\centering
\small
\begin{tabular}{lc}
\toprule
Algorithm & Cumulative regret \\
\midrule
\textbf{Nonparametric TS} & \textbf{100.0 $\pm$ 188.9} (0\%) \\
LinearGaussian TS & 142.1 $\pm$ 200.9 (42.1\%) \\
NeuralLinear & 196.8 $\pm$ 273.6 (96.9\%) \\
NeuralBootstrapped & 214.4 $\pm$ 546.0 (114.5\%) \\
NeuralRMS & 218.2 $\pm$ 556.3 (118.3\%) \\
NeuralDropoutRMS & 196.5 $\pm$ 510.8 (96.5\%) \\
NeuralParamNoise & 130.7 $\pm$ 197.6 (30.8\%) \\
MultitaskGP & 532.7 $\pm$ 310.5 (432.9\%) \\
BNNVariationalGaussian & 323.2 $\pm$ 691.8 (223.3\%) \\
BNNAlphaDiv & 611.6 $\pm$ 187.6 (511.8\%) \\
\bottomrule
\end{tabular}
\end{table}

\begin{figure}[!h]
\centering
\includegraphics[width=0.5\linewidth]{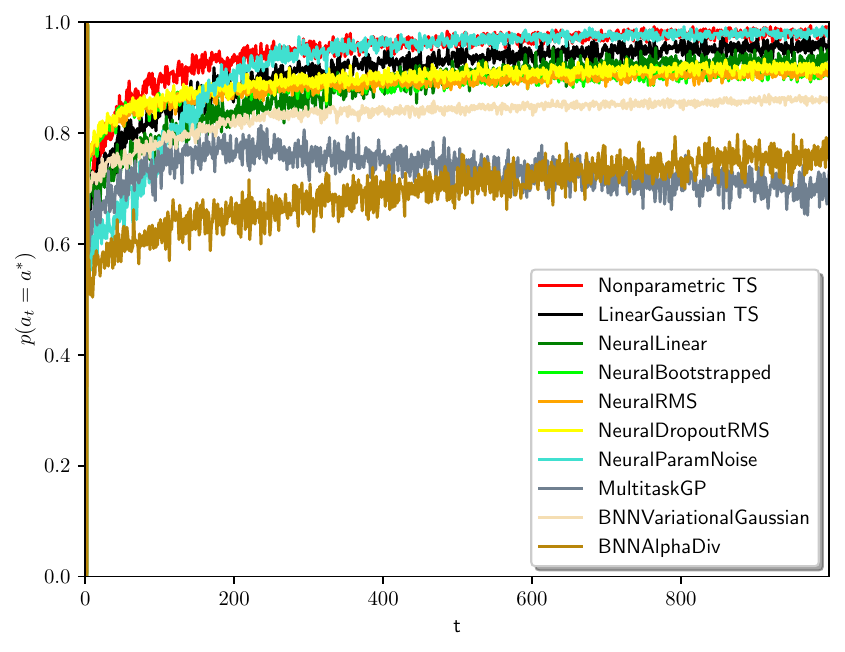}
\caption{Averaged empirical frequency of playing the optimal arm, outlier-contaminated bandit.}
\end{figure}

\subsection{Exponential-reward bandit: full results}
\label{asec:exponential_baselines}

Per-arm rewards $y\geq 0$ are exponentially distributed with context-dependent rate $\lambda_a = \theta_a^\top x_t$, with $x \sim \U{0,1}^2$ and $\theta_a = a \cdot [1,1]^\top$; rewards are thus contextual in both mean and variance. No Oracle TS implementation is available for this reward family.

\begin{table}[!h]
\caption{$R_T$ (mean $\pm$ std) at $T=1000$. Parentheses: relative increase over \texttt{Nonparametric TS}.}
\label{tab:exponential_regret_full}
\centering
\small
\begin{tabular}{lc}
\toprule
Algorithm & Cumulative regret \\
\midrule
\textbf{Nonparametric TS} & \textbf{67.9 $\pm$ 119.4} (0\%) \\
LinearGaussian TS & 116.4 $\pm$ 129.2 (71.5\%) \\
NeuralLinear & 94.2 $\pm$ 125.7 (38.8\%) \\
NeuralBootstrapped & 138.6 $\pm$ 179.9 (104.2\%) \\
NeuralRMS & 143.7 $\pm$ 173.7 (111.8\%) \\
NeuralDropoutRMS & 110.7 $\pm$ 174.0 (63.1\%) \\
NeuralParamNoise & 146.5 $\pm$ 128.5 (115.9\%) \\
MultitaskGP & 197.2 $\pm$ 140.9 (190.5\%) \\
BNNVariationalGaussian & 166.2 $\pm$ 227.6 (144.9\%) \\
BNNAlphaDiv & 153.5 $\pm$ 122.5 (126.1\%) \\
\bottomrule
\end{tabular}
\end{table}

\begin{figure}[!h]
\centering
\includegraphics[width=0.5\linewidth]{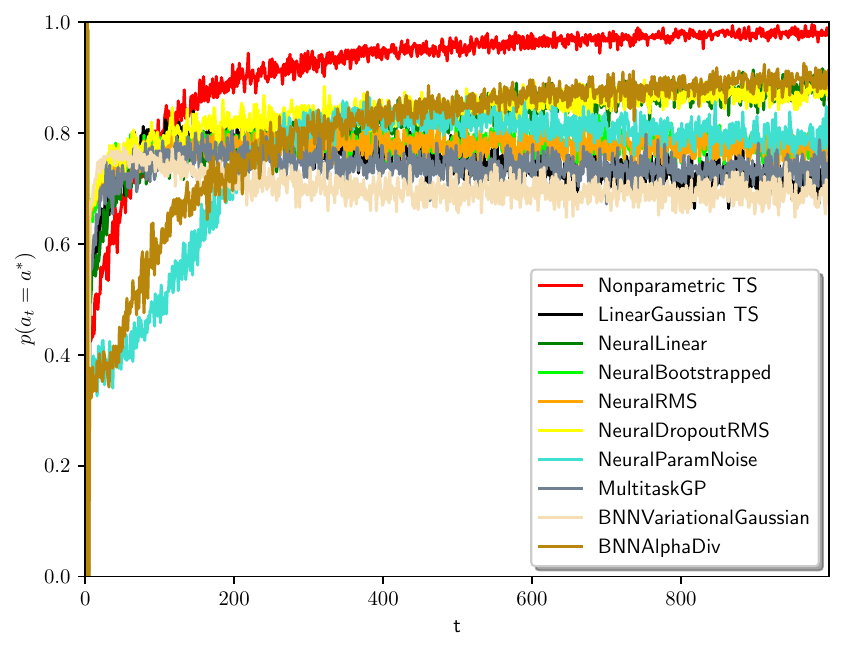}
\caption{Averaged empirical frequency of playing the optimal arm, exponential-reward bandit.}
\end{figure}

\clearpage
\subsection{Gibbs sampler budget: full ablation}
\label{asec:gibbs}

We evaluate the warm-started Gibbs sampler of Section~\ref{ssec:inference} with $Gibbs_{max}\in\{1,5,10\}$, against running to full log-likelihood convergence. In all our experiments, a 0.01\% relative log-likelihood improvement between iterations is typically reached within $Gibbs_{max}\leq10$ iterations, aided by the warm start: at each interaction, the sampler is initialized from the assignments and hyperparameters already inferred for the played arm's previous $n_{t,a}-1$ rewards, rather than from the prior.

\begin{figure}[!h]
\centering
\includegraphics[width=0.48\linewidth]{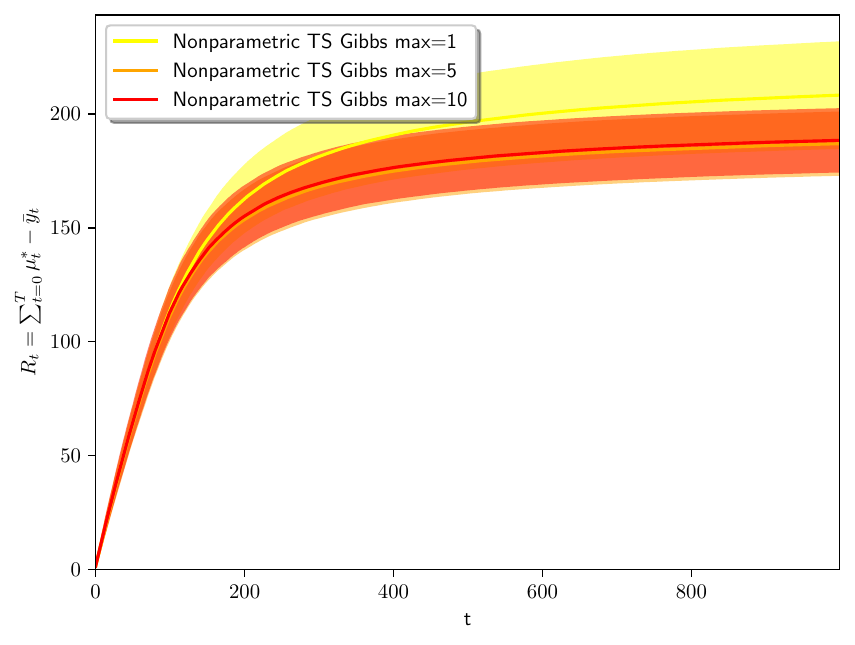}
\includegraphics[width=0.48\linewidth]{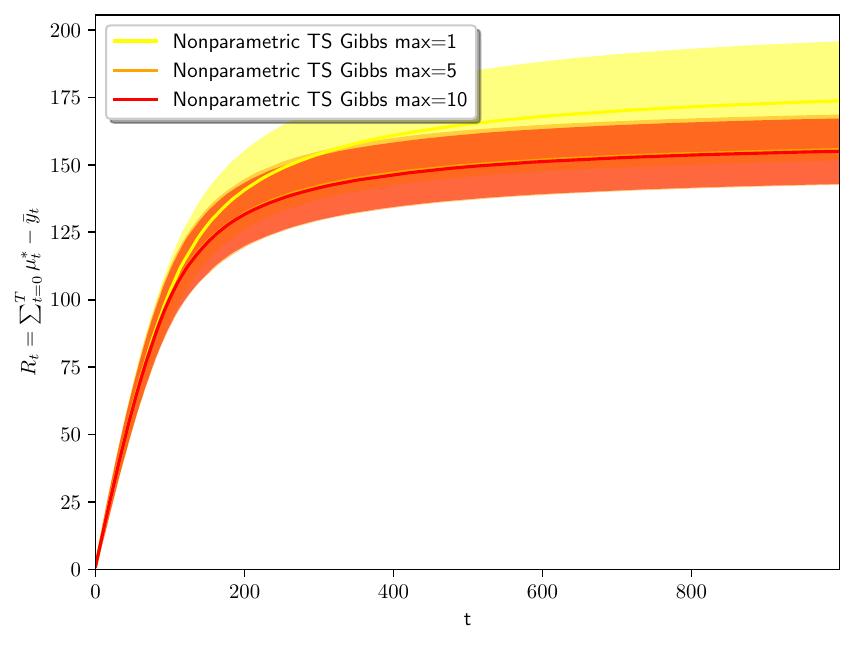}
\caption{Mean $R_t$ (std.\ shaded) for different $Gibbs_{max}$, linear (left) and sparse linear (right) contextual bandits.}
\end{figure}

\begin{figure}[!h]
\centering
\includegraphics[width=0.48\linewidth]{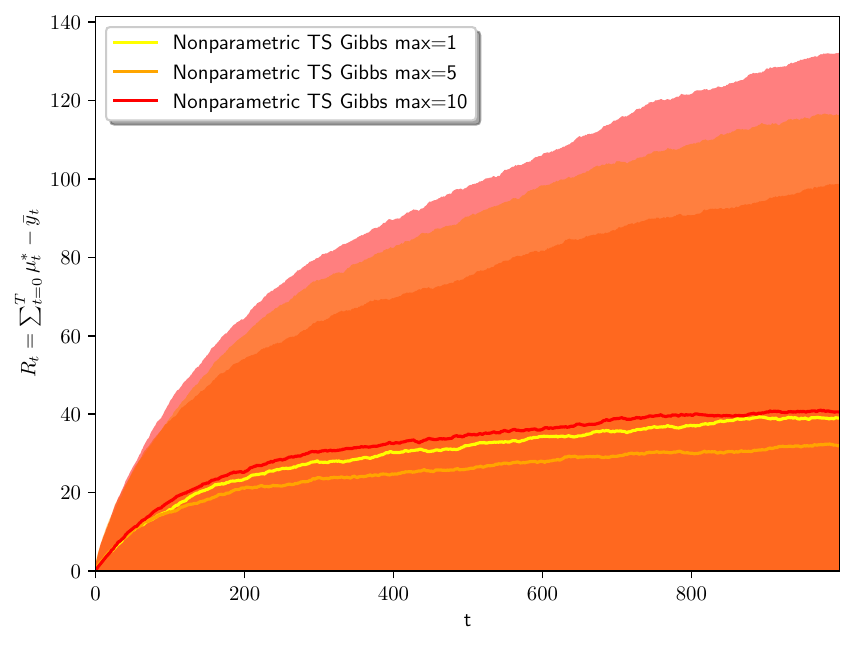}
\includegraphics[width=0.48\linewidth]{figs/linear_gaussian_mixture_unbalanced_gibbsmaxiter/cum_optexpected_regret_std}

\vspace{1ex}
\includegraphics[width=0.48\linewidth]{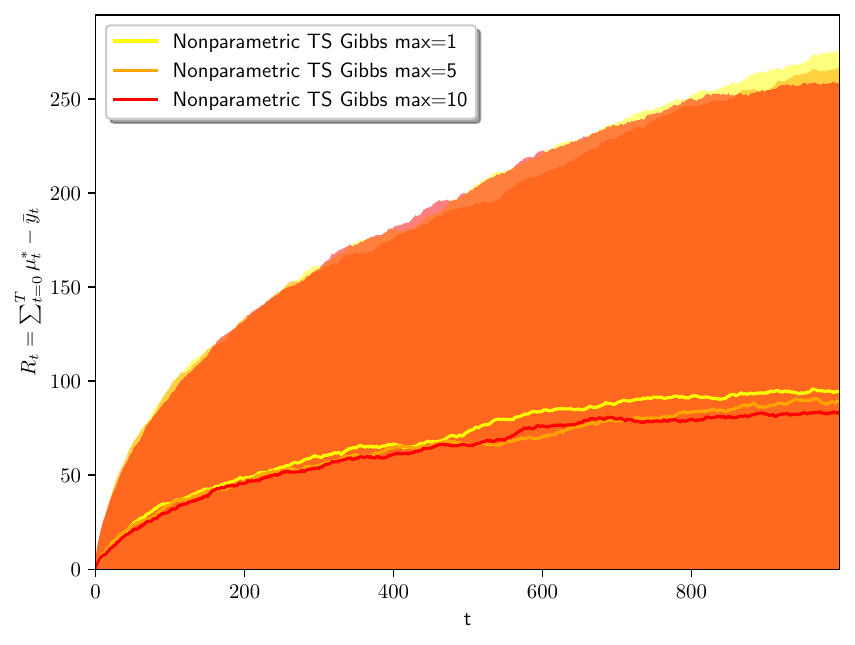}
\includegraphics[width=0.48\linewidth]{figs/exponential_contextual_lambda_gibbsmaxiter/cum_optexpected_regret_std}
\caption{Mean $R_t$ (std.\ shaded) for different $Gibbs_{max}$: Scenario \texttt{A} (top left), Scenario \texttt{B} (top right), outlier-contaminated (bottom left), exponential (bottom right).}
\label{afig:bandit_showdown_gibbsmaxiter}
\end{figure}

Across all scenarios, $Gibbs_{max}=5$ and $Gibbs_{max}=10$ attain comparable regret, while $Gibbs_{max}=1$ measurably increases regret for linear and sparse linear rewards (by about 11\%) and, most clearly, for exponentially distributed rewards (by about 70\%), with no clear difference in the mixture scenarios.
These results suggest that, with the warm start, a few sweeps over an already-informative posterior suffice to incorporate the newest observation well enough to guide the next arm selection, whereas never revisiting past assignments ($Gibbs_{max}=1$) does not.

\clearpage
\subsection{Dirichlet process concentration: sensitivity}
\label{asec:gamma}

The concentration hyperparameter $\gamma_a$ controls how sparse a draw $G_a$ from the Dirichlet process is: as $\gamma_a\to0$, mass concentrates on a single atom; as $\gamma_a\to\infty$, draws approach the (continuous) base measure. We evaluate $\gamma_a\in\{0.01,0.1,1,5,10\}$ across all scenarios of Section~\ref{sec:evaluation}.

\begin{figure}[!h]
\centering
\includegraphics[width=0.48\linewidth]{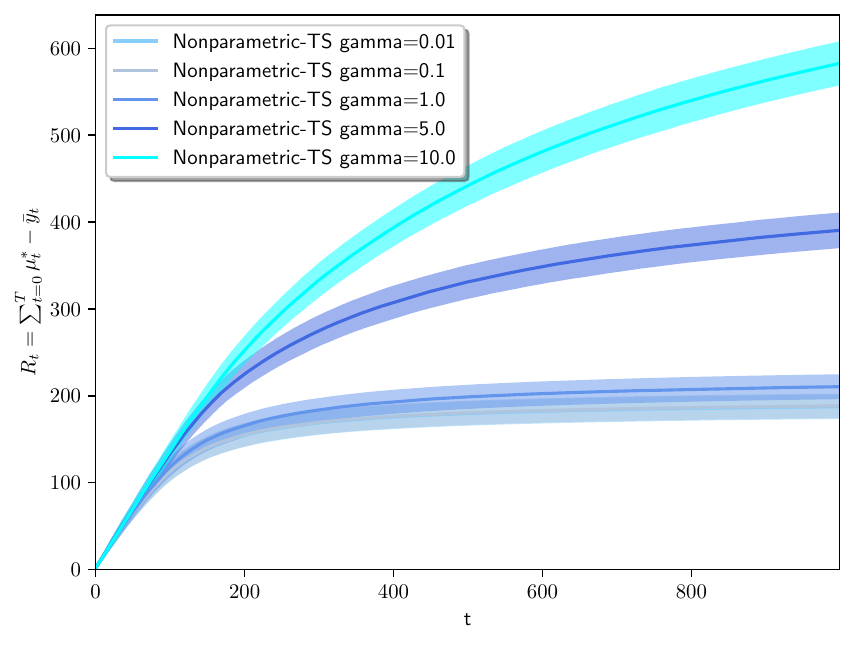}
\includegraphics[width=0.48\linewidth]{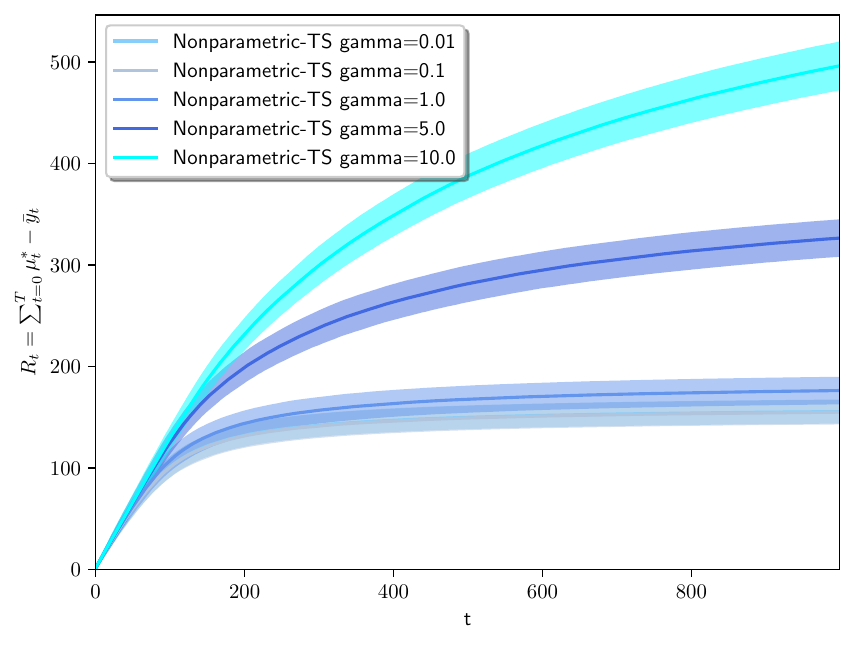}
\caption{Mean $R_t$ (std.\ shaded) for different $\gamma$, linear (left) and sparse linear (right) contextual bandits.}
\end{figure}

\begin{figure}[!h]
\centering
\includegraphics[width=0.48\linewidth]{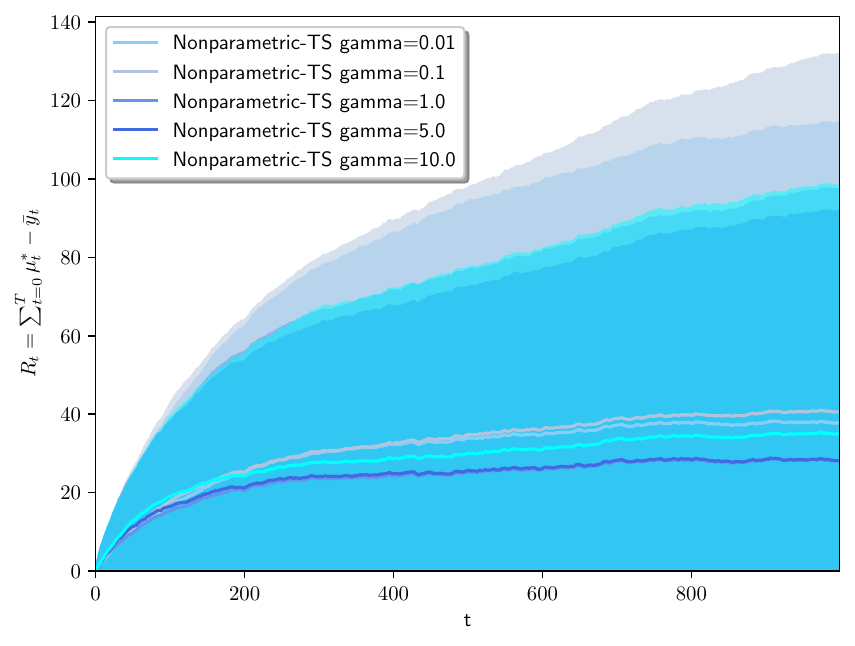}
\includegraphics[width=0.48\linewidth]{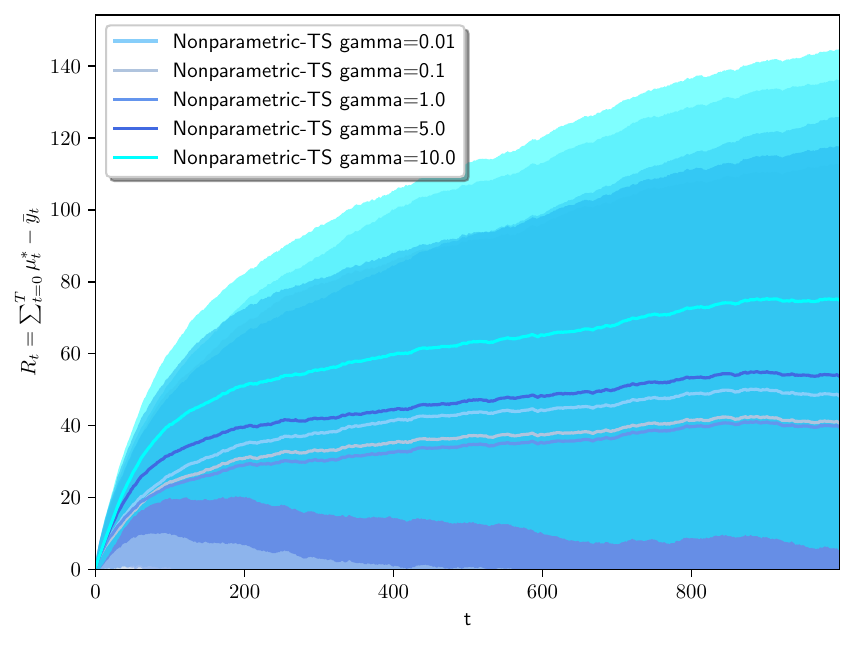}

\vspace{1ex}
\includegraphics[width=0.48\linewidth]{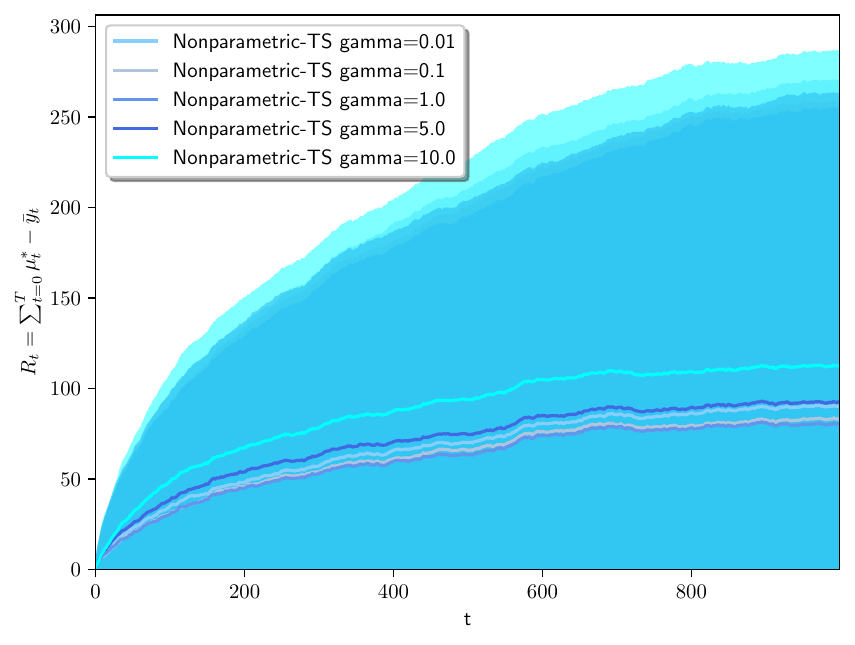}
\includegraphics[width=0.48\linewidth]{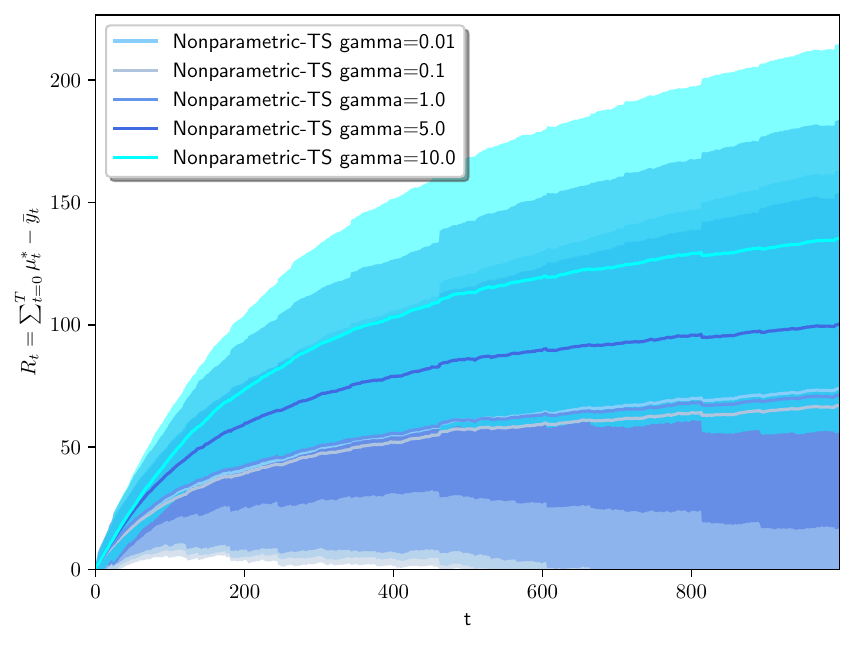}
\caption{Mean $R_t$ (std.\ shaded) for different $\gamma$: Scenario \texttt{A} (top left), Scenario \texttt{B} (top right), outlier-contaminated (bottom left), exponential (bottom right).}
\end{figure}

For unimodal rewards, sparsity-inducing priors ($\gamma_a\leq1$) perform best, as expected, with no significant difference among $\gamma_a\in\{0.01,0.1,1\}$. For mixture-model rewards, a less sparsity-inducing $\gamma_a=1$ is most competitive, while $\gamma_a=0.1$ gives a slight edge for exponentially distributed rewards. Overall, without prior knowledge of reward sparsity, $\gamma_a\in[0.1,1.0]\;\forall a$ gives robust, satisfactory performance across every scenario we studied --- the value $\gamma_a=0.1$ used throughout Section~\ref{sec:evaluation} was not tuned per scenario.

\clearpage
\subsection{Run-times}
\label{asec:runtime}

All experiments were run on CPUs only, as single-core jobs on an internal SLURM cluster
with dual Intel Xeon E5-2650 v4 (2.2 GHz) nodes, requesting up to 64 GB of memory per job.
We report run-times for the experiments of Section~\ref{sec:evaluation}, with two important caveats and limitations in mind.
First, implementations differ: \texttt{Nonparametric TS} is implemented in standard Python (numpy/scipy), while the remaining baselines use the TensorFlow implementation of~\citet{ip-Riquelme2018}; we did not optimize our implementation's efficiency.
Second, all methods (ours and the baselines) are updated at every interaction as a function of the number of observations seen so far, so retraining frequency and duration are confounded with the specific default settings of~\citet{ip-Riquelme2018}, which we did not retune (Section~\ref{sec:discussion}).

\begin{figure}[!h]
\centering
\includegraphics[width=0.48\linewidth]{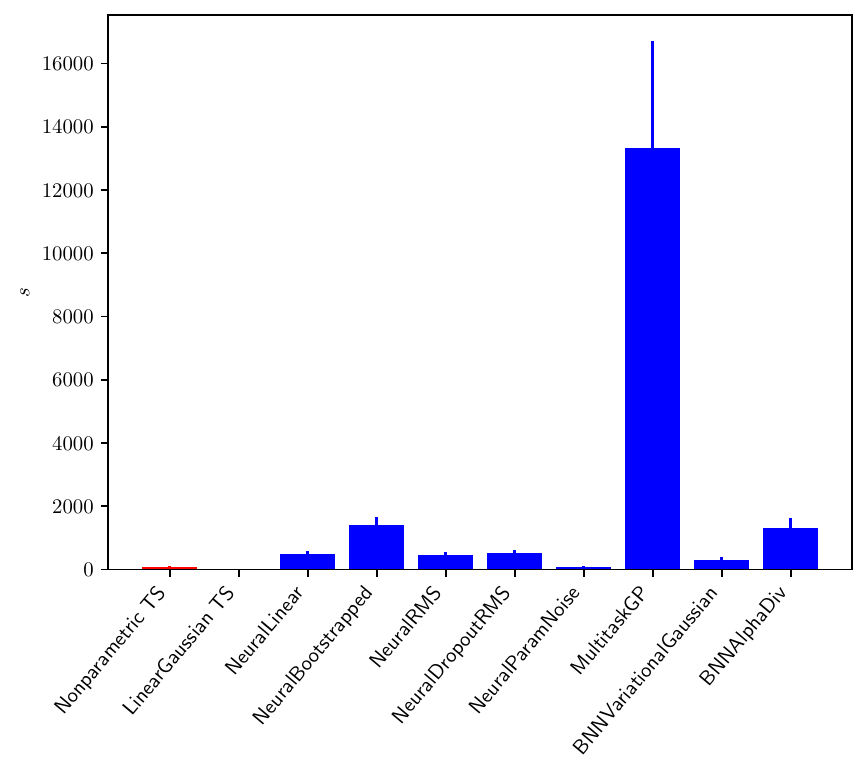}
\includegraphics[width=0.48\linewidth]{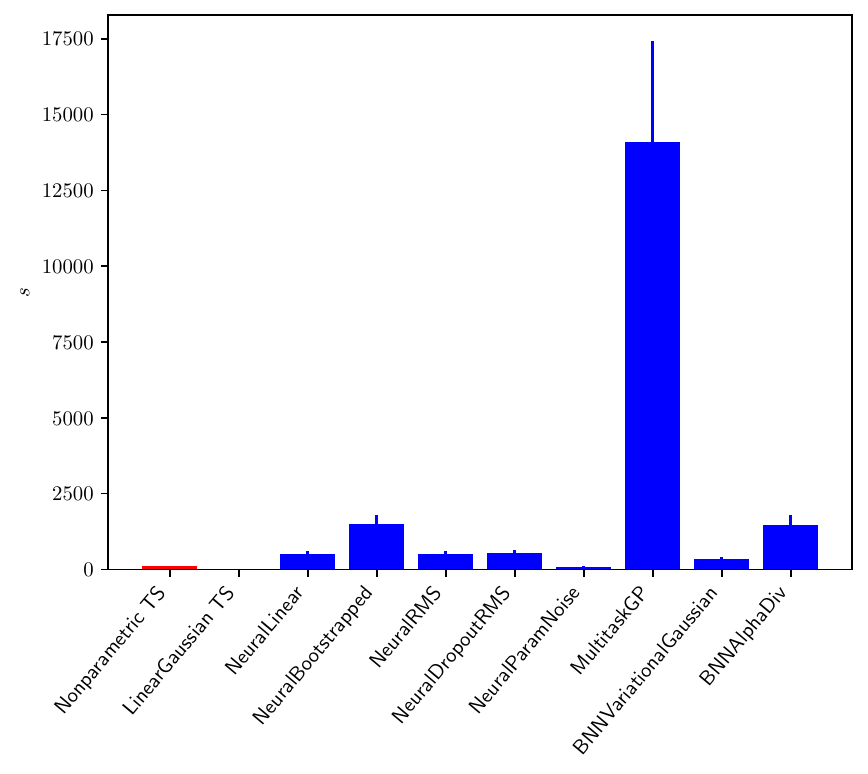}
\caption{Mean run-time in seconds (error bars: one standard deviation), linear (left) and sparse linear (right) contextual bandits.}
\end{figure}

\begin{figure}[!h]
\centering
\includegraphics[width=0.48\linewidth]{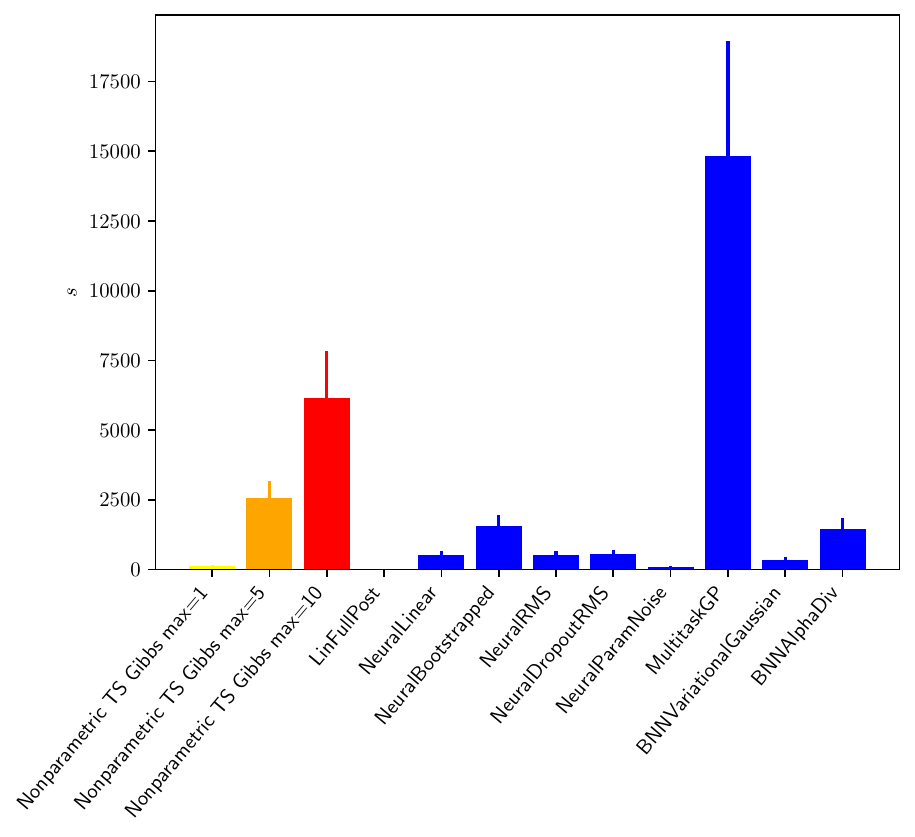}
\includegraphics[width=0.48\linewidth]{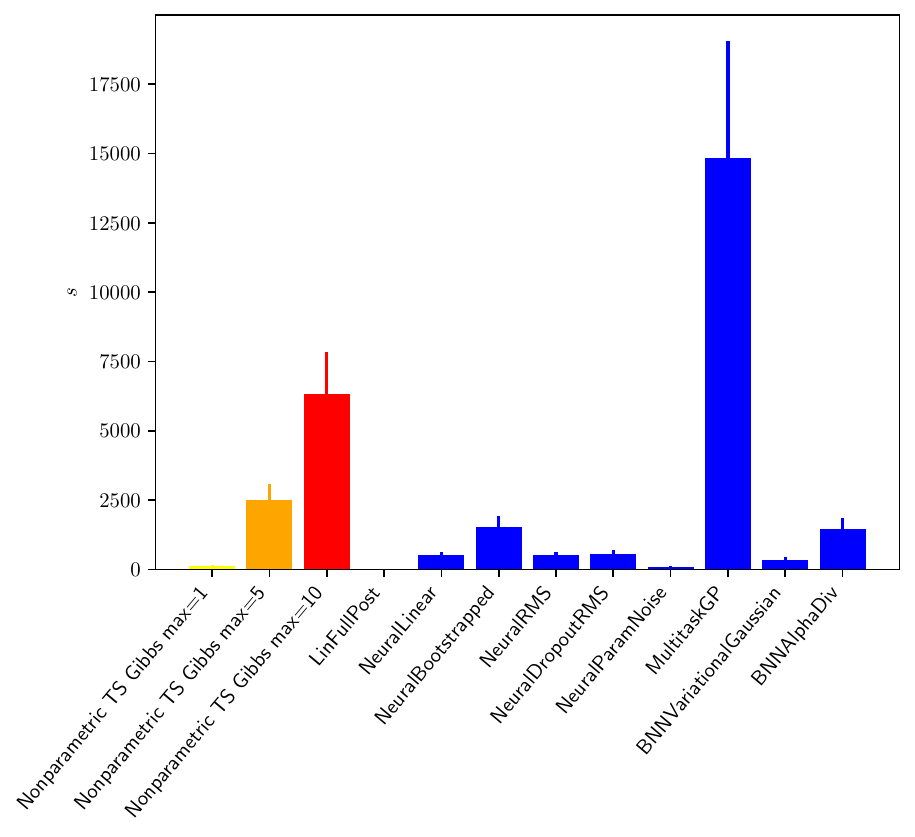}

\vspace{1ex}
\includegraphics[width=0.48\linewidth]{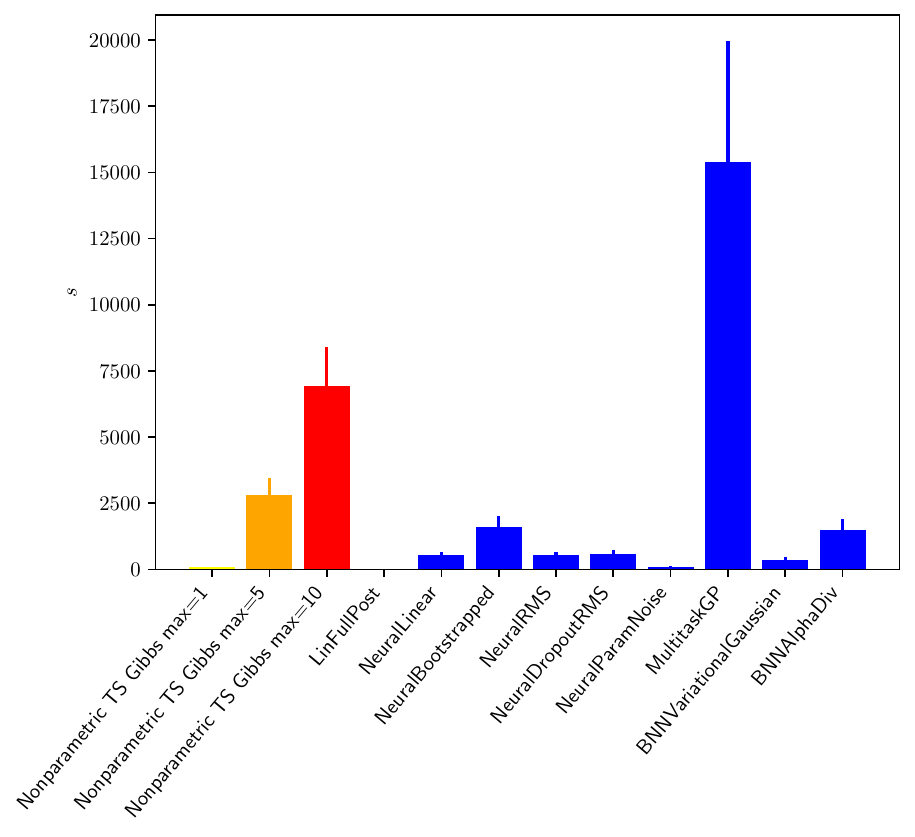}
\includegraphics[width=0.48\linewidth]{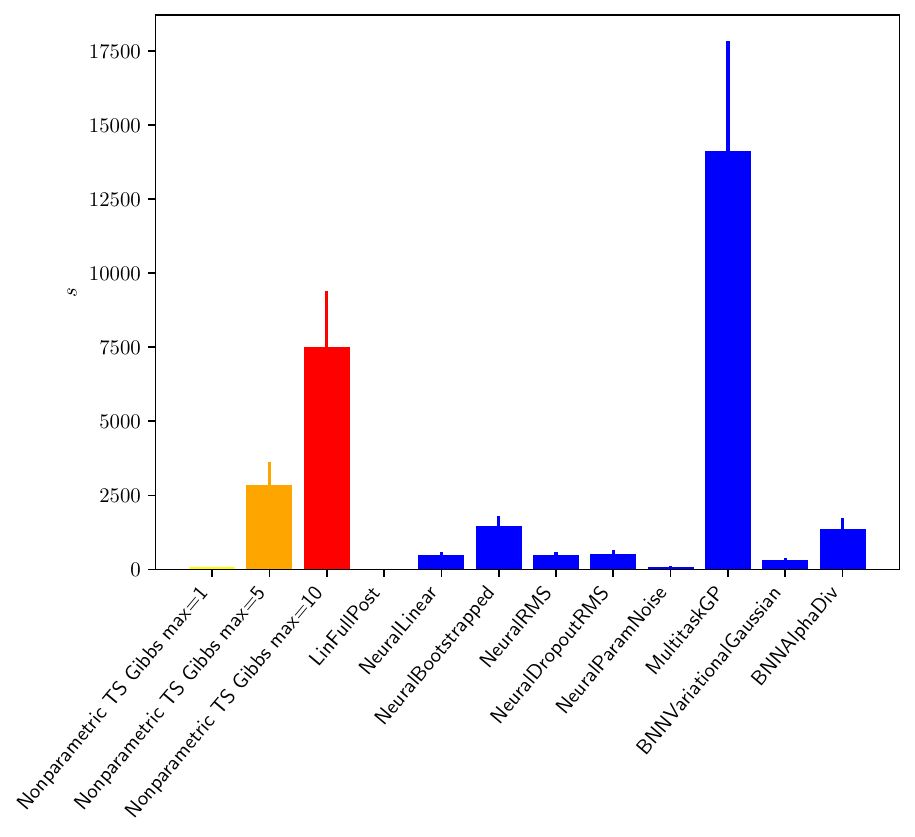}
\caption{Mean run-time in seconds (error bars: one standard deviation): Scenario \texttt{A} (top left), Scenario \texttt{B} (top right), outlier-contaminated (bottom left), exponential (bottom right).}
\end{figure}

\texttt{LinearGaussian TS}, with sequential conjugate updates, is the fastest method in every scenario. \texttt{Nonparametric TS}'s run-time grows with the complexity of the underlying reward model --- low in the linear Gaussian scenarios, higher in the mixture, outlier-contaminated and exponential scenarios --- but, as shown in Appendix~\ref{asec:gibbs}, capping the number of Gibbs iterations substantially reduces this cost while retaining regret performance for $Gibbs_{max}\in[5,10]$: \eg $Gibbs_{max}=5$ is about 2.5$\times$ faster than $Gibbs_{max}=10$ in the mixture scenarios, with comparable regret. This is analogous to the practice of limiting per-iteration training epochs for the neural-network baselines.

\clearpage
\subsection{Real-data bandit: precision oncology}
\label{asec:oncology}

As a real-world stress test, we frame a precision oncology dataset --- genomic features of cancer cell-lines and their in vitro responses to therapeutic protocols, from the publicly available \href{https://github.com/NiklasTR/oncoassign}{precision oncology repository} (which provides no license; hence, we do not redistribute the data, but provide a script that downloads it) --- as a contextual MAB: what drug should be chosen for a given cell-line so that outcomes are maximized. We use log-transformed, median-centered drug response scores (reversed, so higher is better) as rewards, a 20-dimensional UMAP projection of genomic features~\citep{j-Rindtorff2019} concatenated with a one-hot encoding of the dataset's suggested therapeutic protocol (which coincides with the best therapy for only 15\% of cell-lines) as context, and the seven candidate therapies (afatinib, cisplatin, dabrafenib, olaparib, palbociclib, trametinib, vismodegib) as arms. Contexts are sampled with replacement from the 896 cell-lines, and rewards are deterministic given the cell-line and therapy.

\begin{figure}[!h]
\centering
\includegraphics[width=0.48\linewidth]{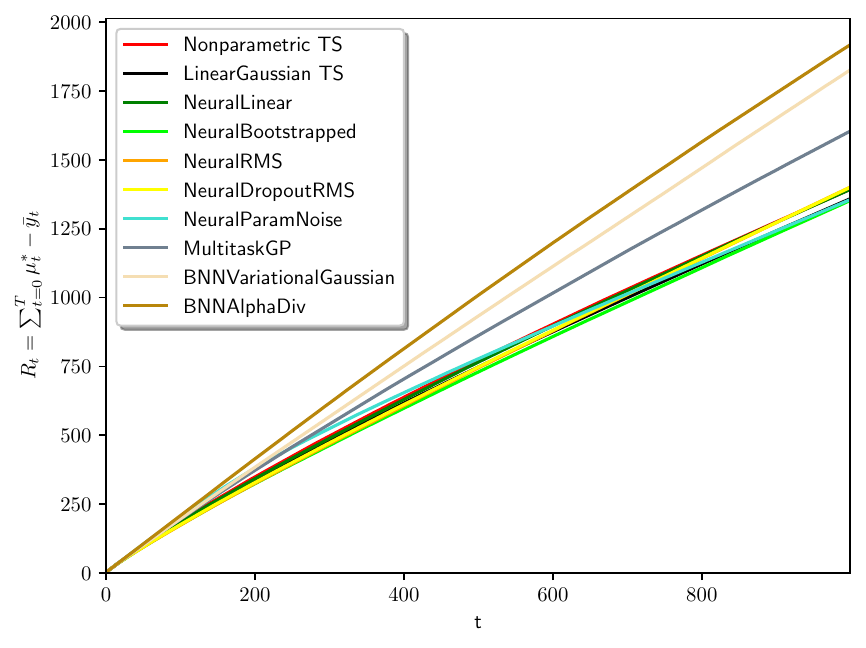}
\includegraphics[width=0.48\linewidth]{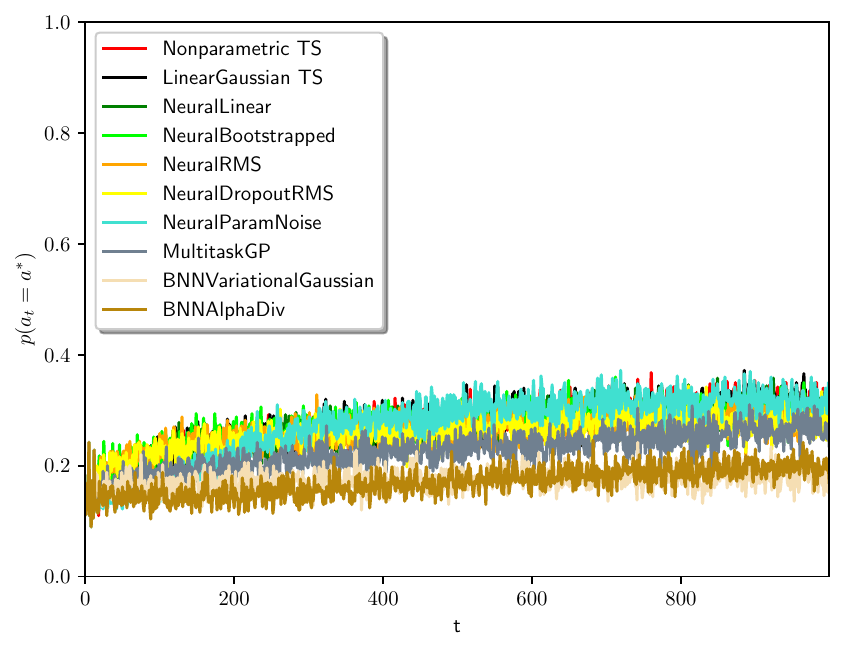}
\caption{Mean $R_t$ (left) and empirical frequency of playing the optimal arm (right), precision oncology bandit.}
\label{afig:precision_oncology}
\end{figure}

This is a considerably harder task than our simulated scenarios: as Figure~\ref{afig:precision_oncology} shows, no method --- including \texttt{Nonparametric TS} --- reliably identifies the best arm, and among the best-performing methods, the variability of cumulative regret across realizations is much larger than the differences in their means (Table~\ref{tab:precision_oncology}).

\texttt{LinearGaussian TS}, \texttt{NeuralBootstrapped} and \texttt{NeuralParamNoise} attain slightly lower mean regret than \texttt{Nonparametric TS} (by 2.5--2.9\%),
which performs on par with \texttt{NeuralLinear}, \texttt{NeuralRMS} and \texttt{NeuralDropoutRMS};
the bootstrapped, RMS and dropout baselines do so with 2--2.6$\times$ higher volatility, consistent with the volatility pattern observed throughout Section~\ref{sec:evaluation}.
\texttt{MultitaskGP} and the BNN-based baselines perform markedly worse (15--38\% higher regret).

We report this result for completeness and as an honest stress test, not as a win for any method:
the differences among the best-performing methods are small in practical terms, see Section~\ref{sec:discussion}.

\begin{table}[!h]
\caption{$R_T$ (mean $\pm$ std) at $T=1000$, precision oncology bandit. Parentheses: relative change vs.\ \texttt{Nonparametric TS}.}
\label{tab:precision_oncology}
\vspace{-2ex}
\centering
\small
\begin{tabular}{lc}
\toprule
Algorithm & Cumulative regret \\
\midrule
\textbf{Nonparametric TS} & \textbf{1392.8 $\pm$ 66.0} (0\%) \\
LinearGaussian TS & 1358.6 $\pm$ 66.2 ($-$2.5\%) \\
NeuralLinear & 1392.5 $\pm$ 68.1 (0.0\%) \\
NeuralBootstrapped & 1352.0 $\pm$ 139.2 ($-$2.9\%) \\
NeuralRMS & 1400.8 $\pm$ 170.8 (0.6\%) \\
NeuralDropoutRMS & 1400.1 $\pm$ 167.6 (0.5\%) \\
NeuralParamNoise & 1355.2 $\pm$ 86.0 ($-$2.7\%) \\
MultitaskGP & 1604.1 $\pm$ 114.5 (15.2\%) \\
BNNVariationalGaussian & 1826.2 $\pm$ 207.7 (31.1\%) \\
BNNAlphaDiv & 1917.6 $\pm$ 63.6 (37.7\%) \\
\bottomrule
\end{tabular}
\end{table}

\clearpage
\section{BNP-based Thompson sampling extensions}
\label{asec:nonparametric_ts_extensions}
\subsection{Pitman-Yor mixture model extension}
\label{asec:pitman_yor}

The model and Thompson sampling algorithm of Section~\ref{sec:method} generalize directly to Pitman-Yor (PY) process priors~\citep{j-pitmanyor1997}, which allow for power-law tail behavior beyond what the Dirichlet process affords~\citep{j-Ishwaran2001}.
We present this extension here for completeness; it is not evaluated empirically in this work.

Replacing the DP prior of Equation~\eqref{eq:per_arm_dp} with a per-arm PY process, $G_a \sim PY(\varphi_a|d_a,\gamma_a,G_{a,0})$ with discount parameter $0\leq d_a<1$, the predictive posterior of Equation~\eqref{eq:nonparametric_Gaussian_mixture_detailed} becomes
\begin{align}
Y_{t+1,a} | x_{t+1}, y_{(t_a)}, \varphi_{t,a}, d_a, \gamma_a, G_{a,0} \sim
& \sum_{k=1}^{K_{t,a}} \frac{n_{t,a,k}-d_a}{n_{t,a}+\gamma_a} \N{Y|x_{t+1}^\top w_{a,k}^*, \sigma_{a,k}^{*2}} \nonumber \\
& + \frac{\gamma_a+K_{t,a}d_a}{n_{t,a}+\gamma_a} \N{Y|x_{t+1}^\top w_{a,k_{new}}, \sigma_{a,k_{new}}^2} \;,
\label{eq:py_predictive}
\end{align}
with a correspondingly modified expected reward,
\begin{align}
\mu_{t,a}(x_t,\varphitilde_a^{(t)}) &= \sum_{k=1}^{K_{t,a}} \frac{n_{t,a,k}-d_a}{n_{t,a}+\gamma_a}\left(x_t^\top \tilde{w}_{a,k}^{(t)*}\right) + \frac{\gamma_a+K_{t,a}d_a}{n_{t,a}+\gamma_a}\left(x_t^\top \tilde{w}_{a,k_{new}}^{(t)}\right) \;.
\label{eq:py_expected_reward}
\end{align}

The discount parameter $d_a$ gives the PY process additional control over tail behavior relative to the DP (recovered as the special case $d_a=0$), whose number of distinct components grows only logarithmically with the number of observations; we leave an empirical study of this extension, and of when its added flexibility is worth the extra hyperparameter, for future work.

\subsection{Hierarchical mixture model extension}
\label{asec:hierarchical}

Section~\ref{ssec:nonparametric_rewards} models each arm's reward distribution with an \emph{independent} BNP mixture. 

An alternative BNP bandit model is a \emph{hierarchical} Dirichlet (or Pitman-Yor) process mixture, in which every arm's random reward distribution $G_a$ is drawn from a common, shared base measure $G_0$~\citep{j-Teh2010}. We record this extension here for completeness; it is not evaluated empirically in this work.

The generative process of a hierarchical Pitman-Yor mixture model, illustrated in Figure~\ref{afig:pgm_nonparametric_bandit_hierarchical}, is
\begin{align}
G_0 &\sim PY(\varphi|d_0,\gamma_0,H) \;, \\
G_a &\sim PY(\varphi_a|d_a,\gamma_a,G_0) \;, \; \forall a\in\A \;, \\
\varphi_{t,a} &\sim G_a \;, \\
Y_{t,a} &\sim p(Y|x_t,\varphi_{t,a}) = \N{Y|x_t^\top w_{t,a},\sigma_{t,a}^2} \;,
\end{align}
where the hierarchical Dirichlet process is the special case $d_0=d_a=0$.

\begin{figure}[!h]
\centering
\begin{tikzpicture}
	\node[obs] (y-t) {$y_{t}$};
	\node[latent, above=1.0 of y-t, xshift=0cm] (varphi-ta) {$\varphi_{t,a}$};
	\node[latent, above=0.5 of varphi-ta, xshift=0cm] (G-a) {$G_{a}$};
	\node[latent, above=1.5 of G-a, xshift=0cm] (G-0) {$G_{0}$};
	\node[latent, left=1 of y-t] (a-t) {$a_t$};
	\node[latent, below=0.5 of y-t]  (x-t) {$x_t$};
	
	\node[const, above=0.5 of G-a, xshift=-1.0cm] (d-a) {$d_{a}$} ;
	\node[const, above=0.5 of G-a, xshift=1.0cm] (gamma-a) {$\gamma_{a}$} ;
	\node[const, above=0.5 of G-0, xshift=-1.0cm] (d-0) {$d_{0}$} ;
	\node[const, above=0.5 of G-0, xshift=1.0cm] (gamma-0) {$\gamma_{0}$} ;
	\node[const, above=0.5 of G-0, xshift=0.0cm]  (H) {$H$} ;
	
	\edge {gamma-0,H} {G-0} ;
	\edge {d-0,H} {G-0} ;
	\edge {gamma-a,G-0} {G-a} ;
	\edge {d-a,G-0} {G-a} ;
	\edge {G-a} {varphi-ta} ;
	\edge {varphi-ta,x-t,a-t} {y-t} ;
	
	\plate {t} {
		(a-t)(x-t)(y-t)
		(varphi-ta) %
		} {$t$} ;
	\plate {a}{
		(d-a)(gamma-a)(G-a0) %
		(G-a) %
		(varphi-ta) %
	} {$A$} ;
\end{tikzpicture}
\caption{Graphical model of the hierarchical mixture bandit.}
\label{afig:pgm_nonparametric_bandit_hierarchical}
\end{figure}

Because every arm's distribution is drawn from the same shared $G_0$, rewards observed from arm $a$ inform the posterior over other arms' distributions $a'\neq a$ as well.
Namely, due to the clustering property of the PY (or DP) process, the model defines a global mixture with $K$ distinct atoms, shared across arms with only the mixing proportions varying per arm.

This is a strictly different modeling assumption from Section~\ref{ssec:nonparametric_rewards}: information sharing across arms is gained at the cost of requiring every arm to belong to the same family of reward distributions, which need not hold in general (\eg it does not hold in our \texttt{Scenario B}, Eq.~\eqref{eq:scenario_mixture_B}).
Inference would follow analogously, via a collapsed Gibbs sampler over both local (per-arm) and global mixture-component assignments; we omit the derivation here, as this extension is not evaluated empirically in this work.

\end{document}